\documentclass[table]{article}
\usepackage[preprint]{neurips_2025}
\usepackage{longtable}
\usepackage[most]{tcolorbox}
\usepackage{lipsum} 
\usepackage{setspace} 
\usepackage{caption}
\usepackage{xcolor}
\definecolor{Gray}{gray}{0.93}
\usepackage{multirow}
\usepackage{amsmath}
\usepackage{graphicx}
\usepackage{changepage}
\usepackage[utf8]{inputenc} 
\usepackage[T1]{fontenc}    
\usepackage{hyperref}       
\usepackage{url}            
\usepackage{booktabs}
\usepackage{amsfonts}       
\usepackage{nicefrac}       
\usepackage{microtype}      
\usepackage{subcaption}
\usepackage{pifont} 
\usepackage{colortbl} 
\definecolor{LightRed}{RGB}{255, 200, 200}
\definecolor{LightBlue}{RGB}{200, 220, 255}

\title{Q-Ponder: A Unified Training Pipeline for Reasoning-based Visual Quality Assessment}

\author{Zhuoxuan Cai$^{1,2}$\footnotemark[1] \quad Jian Zhang$^{{2}}$\footnotemark[3] \quad Xinbin Yuan$^2$ \quad Peng-Tao Jiang$^2$ \quad Wenxiang Chen$^1$ \\ \textbf{Bowen Tang}$^{2}$ \quad \textbf{Lujian Yao}$^2$ \quad \textbf{Qiyuan Wang}$^1$ \quad \textbf{Jinwei Chen}$^2$ \quad 
 \textbf{Bo Li}$^2$\footnotemark[2]  \\ \\
$^1$Fudan University \quad $^2$vivo Mobile Communication Co., Ltd   \\
 \textit{\normalsize {zxcai23}@m.fudan.edu.cn, libra@vivo.com} \\  
 {\normalsize Project page: \url{https://vivocameraresearch.github.io/qponder}}
 }

\begin{document}
\footnotetext[1]{This work was done during Zhuoxuan Cai’s internship at vivo.}
\footnotetext[2]{Corresponding authors.}
\footnotetext[3]{Project lead.}
\maketitle

\begin{abstract}
   Recent studies demonstrate that multimodal large language models (MLLMs) can proficiently evaluate visual quality through interpretable assessments. However, existing approaches typically treat quality scoring and reasoning descriptions as separate tasks with disjoint optimization objectives, leading to a trade-off: models adept at quality reasoning descriptions struggle with precise score regression, while score-focused models lack interpretability. This limitation hinders the full potential of MLLMs in visual quality assessment, where accuracy and interpretability should be mutually reinforcing. To address this, we propose a unified two-stage training framework comprising a cold-start stage and a reinforcement learning-based fine-tuning stage. Specifically, in the first stage, we distill high-quality data from a teacher model through expert-designed prompts, initializing reasoning capabilities via cross-entropy loss supervision. In the second stage, we introduce a novel reward with Group Relative Policy Optimization (GRPO) to jointly optimize scoring accuracy and reasoning consistency. We designate the models derived from these two stages as \textbf{Q-Ponder-CI} and \textbf{Q-Ponder}. Extensive experiments show that Q-Ponder achieves state-of-the-art (SOTA) performance on quality score regression benchmarks, delivering up to 6.5\% higher SRCC on cross-domain datasets. Furthermore, Q-Ponder significantly outperforms description-based SOTA models, including its teacher model Qwen-2.5-VL-72B, particularly in description accuracy and reasonableness, demonstrating the generalization potential over diverse tasks.
\end{abstract}

\section{Introduction}
Evaluating image quality by emulating human reasoning has long been a frontier topic in computer vision.  
A mature image quality assessment~(IQA) system can effectively emulate human perception to provide quality feedback for tasks such as image enhancement~\cite{chen2024prompt, zhou2022quality, zhu2024intelligent, zheng2021learning}, image generation~\cite{chen2024study, yu2024sf}, and computational photography~\cite{cai2024phocolens, fang2020perceptual}, thereby supporting downstream image processing and optimization. Traditional handcrafted approaches predict image quality from multi‐perspective statistical cues and human prior knowledge\cite{hore2010image, zhang2011fsim}, but their inability to model high‐level semantics keeps performance bounded\cite{bosse2017deep}. Deep learning-based methods alleviate this limitation by automatically learning low‐ and high‐level semantic features and regressing quality scores\cite{talebi2018nima, zhang2018unreasonable}, yet their black‐box nature and the bias/cost of human‐curated data hinder interpretability and out‐of‐domain robustness.

\begin{figure}[!t]
    \centering
    \includegraphics[width=\textwidth]{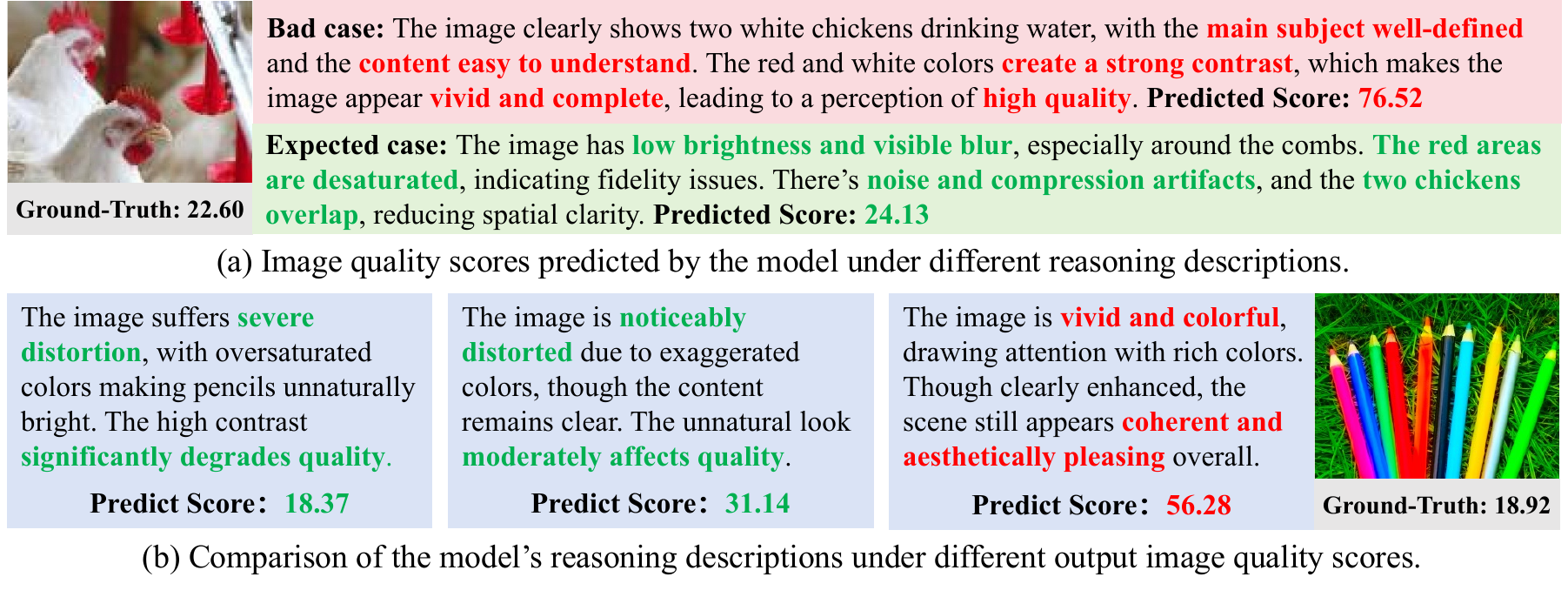}
    \caption{\textbf{Illustrations of the motivation.} (a) A comprehensive, reasonable, and accurate "expert-level" reasoning process helps the model regress precise quality scores, while maintaining a certain level of robustness on out-of-distribution data during training.
(b) As the predicted score approaches the ground truth, the precision of the reasoning descriptions improves, indicating that it is possible to refine the model’s reasoning process while simultaneously encouraging accurate score regression.}
    \label{fig1}
    \vspace{-15pt}
\end{figure}

Multimodal large language models~(MLLMs) have demonstrated remarkable capabilities in general vision-language tasks through large‐scale cross‐modal pretraining and rich linguistic generation~\cite{zhang2023gpt, ye2024mplug, bai2025qwen2, chiang2023vicuna, team2023internlm, alayrac2022flamingo}.
Inspired by these advances, researchers have recently begun adapting MLLMs to image quality assessment\cite{cao2023comprehensive, you2024depicting}, uncovering their potential for robust cross-domain generalization and interpretable evaluations.
Current studies on MLLM-based IQA generally fall into two distinct directions:

One line of methods aims to \textbf{directly regress image quality scores with high precision}, striving to produce quantitative assessments that align closely with human annotations. Representative approaches include \textbf{Q-Align}~\cite{wu2023q}, \textbf{DeQA}~\cite{you2025teaching}, and \textbf{Compare2Score}~\cite{zhu2024adaptive}, which achieve this goal via token-level supervision or pairwise preference modeling. Q-Align and DeQA discretize continuous quality scores into language tokens and apply standard language modeling objectives to predict the next token, thereby reframing quality assessment as a token prediction task. In contrast to token-based approaches, Compare2Score adopts a preference modeling strategy: it computes pairwise preference probabilities between a target image and a set of anchor images, constructs a probability matrix, and applies maximum a posteriori (MAP) estimation to infer the final quality score. These methods leverage the strengths of MLLMs in token prediction and preference modeling, producing quality assessments that closely match human-labeled scores and achieving state-of-the-art (SOTA) performance on multiple benchmark datasets. However, their reliance on discretized numerical labels or implicit preference signals at the token level sacrifices interpretability and tends to cause overfitting to shallow numerical patterns, thereby undermining the inherent multimodal reasoning capabilities of MLLMs. As a result, such models often fail to provide meaningful explanations for their predictions and exhibit limited generalization to novel domains~\cite{zhong2024causal, wang2025mp}.

Another line of research explores MLLMs’ capability to \textbf{produce reasoning-based descriptions that explain quality judgments in IQA tasks}. Representative models such as \textbf{DepictQA}~\cite{you2024depicting} and \textbf{Co-Instruct}~\cite{wu2024towards} are trained on large-scale corpora that pair images with human-annotated descriptions, comparisons, and justifications.
These models excel at generating insightful and detailed explanations of quality assessments, capable of describing subtle visual artifacts or stylistic degradations, giving them a natural advantage in certain perception-oriented IQA tasks. However, this descriptive strength often comes at the expense of numerical precision: their outputs tend to lack consistency in scoring and may fail to capture the fine-grained quality distinctions crucial for practical image quality assessment. Moreover, these methods often neglect the latent information embedded in the original score annotations. As a result, they suffer from poor data efficiency and cannot scale effectively due to the high cost of human-curated supervision.

\vspace{-3.5mm}
\paragraph{Motivation.}
Our motivation lies in unifying the long-separated optimization objectives of score prediction and reasoning description tasks, emphasizing their complementary roles. First, we argue that a structured and expert-level reasoning process improves score alignment, as it encourages the model to incorporate interpretable distortion analysis to guide its quality predictions. As illustrated in Fig.~\ref{fig1}(a), when equipped with long-chain reasoning ability, the model can leverage diverse visual cues and distortion patterns to generate more consistent and accurate quality scores. Second, we propose that applying a result-based reward mechanism over the complete output—including both the reasoning and the predicted score—facilitates joint optimization: while aligning numerical scores, the model can simultaneously refine the reasonableness and accuracy of its reasoning without explicit supervision for the reasoning process. As shown in Fig.~\ref{fig1}(b), as score predictions become more accurate, the model’s judgment of distortion types and severity also becomes more reasonable, with reduced hallucinations. We empirically validate this hypothesis in subsequent experiments.

Bridging this gap calls for a unified framework that can simultaneously regress accurate scores and generate coherent, causal explanations. Such a framework maximizes the utility of available data and fully realizes the multimodal capabilities of MLLMs in practical quality evaluation scenarios. Before delving into the technical details, we summarize our key contributions as follows:

\vspace{-2mm}
\begin{itemize}
\item[$\bullet$] [\textbf{Reasoning Distillation Template.}] We propose a heuristic-filtered reasoning distillation method that efficiently captures the intrinsic logic between images and their associated quality labels, without incurring heavy manual annotation costs. By incorporating original distortion types and quality levels into the distillation process, and coupling them with a set of expert-designed prompting templates, our method generates factually grounded initial reasoning paths that significantly enhance the model’s descriptive reasoning capability.

\item[$\bullet$] [\textbf{Reinforcement Learning Fine-tuning Strategy.}] Building upon the cold-start initialization, we introduce a reinforcement learning strategy to finely optimize both the reasoning process and final predictions. By applying a novel reward function over the complete output, the model improves its score accuracy while reducing hallucinations in reasoning, resulting in more reasonable and accurate quality assessments.

\item[$\bullet$] [\textbf{Unified Two-Stage Training Pipeline.}] Leveraging the above techniques, we develop a unified two-stage training pipeline that jointly optimizes quality score regression and reasoning descriptions. Based on this pipeline, we train and release two models: Q-Ponder-CI and Q-Ponder. Extensive experiments demonstrate that Q-Ponder achieves state-of-the-art performance on both quality scoring and descriptive reasoning tasks, while also exhibiting superior generalization on out-of-distribution (OOD) datasets.
\end{itemize}

\section{Related work}

\subsection{Score-based IQA methods}
Prior to MLLMs, IQA primarily relied on full-reference (FR) and no-reference (NR) methods. FR techniques used handcrafted metrics based on human visual system priors, including PSNR \cite{hore2010image}, SSIM\cite{wang2004image}, FSIM \cite{zhang2011fsim}, and VIF\cite{sheikh2006image}. NR methods such as NIQE \cite{mittal2012making} and BRISQUE \cite{mittal2012no} estimated quality from natural scene statistics \cite{ma2017learning, moorthy2010two, moorthy2011blind, tang2011learning, saad2012blind}. While efficient, these methods lacked robustness to high-level semantic content and complex distortions. Deep learning methods leveraging CNNs \cite{bosse2017deep} and vision transformers \cite{cheon2021perceptual} greatly improved generalization by learning perceptual features from large-scale data. In FR IQA, LPIPS \cite{zhang2018unreasonable} and PieAPP \cite{prashnani2018pieapp} are widely used; in NR IQA, NIMA \cite{talebi2018nima}, HyperIQA \cite{su2020blindly}, and MUSIQ \cite{ke2021musiq} achieve strong performance. Recent methods further incorporate graphs \cite{sun2022graphiqa}, CLIP \cite{wang2023exploring}, continual learning \cite{zhang2022continual}, meta-learning \cite{zhu2020metaiqa}, and multi-task strategies \cite{zhang2023blind, ma2017end}. Nonetheless, current models face notable challenges such as overfitting to in-distribution data and difficulty in capturing meaningful reasoning or attribution patterns.

\subsection{MLLM-based IQA methods}
MLLMs\cite{ye2024mplug, alayrac2022flamingo, chiang2023vicuna, zhang2024vision, li2023blip, liu2023visual, team263218031gpt} integrate vision and language, enabling powerful cross-modal understanding and generation. Recent work has adapted MLLMs to IQA, leveraging their perception capabilities, generalization strength, and interpretability \cite{zhang2024quality, wu2024comprehensive}. Q-Bench \cite{wu2023q, zhang2024q} established standardized evaluation protocols and demonstrated MLLMs' ability to perceive low-level visual attributes. Q-Align \cite{wu2023q}, DeQA \cite{you2025teaching}, and C2Score \cite{zhu2024adaptive} formulate quality prediction as classification, distribution regression, or contrastive learning tasks, improving numerical performance but often sacrificing interpretability. In parallel, Q-Instruct \cite{wu2024q} enhances reasoning over distortion semantics via instruction tuning, while DepictQA\cite{you2024depicting} and its extension DepictQA-Wild \cite{you2024descriptive} generate detailed quality descriptions by training on large-scale explanation corpora. Co-Instruct \cite{wu2024towards} enables fine-grained pairwise comparisons. Despite these advances, descriptive methods often suffer from weak numerical accuracy and high annotation costs. Other efforts include ROI-level analysis \cite{chen2024q}, region-aware scoring \cite{chen2024seagull}, and prompt engineering \cite{chen2024promptiqa} to improve reliability and adaptivity.

\subsection{Reinforcement learning}
Reinforcement learning (RL) optimizes policies by interacting with the environment to maximize cumulative rewards. Foundational breakthroughs such as DQN \cite{mnih2015human} (addressing high-dimensional state spaces), AlphaGo \cite{silver2016mastering} (integrating Monte Carlo Tree Search), and PPO \cite{schulman2017proximal} (stabilizing policy optimization) established the basis of modern RL. With the rise of large language models, RL with human feedback (RLHF) \cite{christiano2017deep, ouyang2022training} introduced preference-based fine-tuning to align model outputs with human intent, and was soon extended to multimodal models \cite{yu2024rlaif, yu2024rlhf, zhang2025mm} for improved safety and generalization. More recently, Group Relative Policy Optimization (GRPO) \cite{shao2024deepseekmath} was introduced in DeepSeek-R1 \cite{guo2025deepseek} to enhance training efficiency through dynamic sampling strategies and inference-time scaling. GRPO has shown strong generalization and structured supervision across diverse vision tasks, including multimodal reasoning \cite{shen2025vlm, sheng2024hybridflow}, visual grounding \cite{liu2025visual}, GUI-based agents \cite{lu2025ui, liu2025infigui}, medical visual QA \cite{pan2025medvlm}, and image generation \cite{guo2025can}. In a contemporaneous effort, Q-Insight\cite{li2025q} apply a purely RL paradigm to IQA. However, its discrete reward formulation limits the model’s capacity for precise quality score regression. Moreover, the approach relies on manually annotated reasoning paths, which not only incur significant labeling costs but also impose rigid output format constraints, restricting its applicability to distortion classification tasks. In contrast, a general-purpose IQA model should be capable of open-ended evaluation of natural image quality and provide targeted enhancement suggestions, rather than relying on predefined, annotation-dependent reasoning patterns.

\section{Methodology}
In this section, we first introduce the necessary background for quantitatively evaluating the scoring and reasoning capabilities of MLLMs (Sec.~\ref{Preliminaries}). We then present our expert-designed prompt templates to guide the teacher model in generating score and reasoning outputs from both low- and high-level perspectives, along with a heuristic filtering strategy to eliminate logically flawed samples and improve distillation data quality (Sec.~\ref{Cold start}). Finally, we introduce a GRPO-based reinforcement fine-tuning strategy to further optimize the model into an expert-level IQA system(Sec.~\ref{Reasoning-oriented reinforcement learning}).

\subsection{Preliminaries}
\label{Preliminaries}
\paragraph{Quantitative Evaluation of Scoring Capabilities.}
We adopt two standard correlation metrics to evaluate MLLMs' ability to predict image quality: the Pearson Linear Correlation Coefficient (PLCC) and the Spearman Rank Order Correlation Coefficient (SRCC). PLCC measures the linear correlation between predicted scores and human ratings, reflecting numerical accuracy, while SRCC assesses the consistency of rank ordering, indicating the model’s ability to preserve relative quality relationships.

\vspace{-2mm}
\paragraph{Quantitative Evaluation of Reasoning Capabilities.}
\label{evalR}
Previous studies on descriptive reasoning in IQA typically focus on specific tasks, such as distortion type classification~\cite{you2024depicting,you2024descriptive,wu2024q} or pairwise comparison~\cite{liu2017rankiqa,wu2024towards}, lacking a unified metric for reasoning descriptions of image quality. Q-Bench~\cite{wu2023q, zhang2024q} is the first to introduce export text annotations and use a LLM as a judge to evaluate the completeness, accuracy, and relevance of textual descriptions in their A2 evaluation task. 

To overcome the high annotation cost and the limitation that LLMs cannot directly perceive image content, we propose an improved evaluation strategy using proprietary MLLMs to assess reasoning descriptions: We prioritize datasets with original distortion labels for evaluation and enrich the semantic content of these labels by categorizing distortion severity into five levels: \textit{extreme, severe, noticeable, moderate}, and \textit{slight}, based on an equal partitioning of the MOS range [0, 1]. Each distorted image is paired with its quality description and the refined severity label. The resulting outputs are then evaluated by a proprietary MLLM across the following dimensions:

\begin{tcolorbox}[
    colback=blue!5,
    colframe=black,
    sharp corners=south,
    enhanced,
    arc=0mm, 
    boxrule=0.8pt,
    left=2mm,
    right=2mm,
    top=1mm,
    bottom=1mm
]

\textbf{Completeness:} Does the response cover all major distortion elements and reflect diverse evaluation perspectives?\par\vspace{2pt}
\textbf{Accuracy:} Does the response correctly identify the dominant distortions and assign a severity level aligned with the ground-truth label?\par\vspace{2pt}
\textbf{Reasonableness:} Is the reasoning logically coherent, free from contradictions or hallucinations, and semantically consistent with the image content?

\end{tcolorbox}

A quality score from 0 to 5 is assigned to each dimension by the judge model. This multi-perspective evaluation enables us to dissect the model’s reasoning chain — completeness measures breadth, accuracy ensures fidelity to the ground-truth, and reasonableness guards against implausible or irrelevant reasoning. We adopt Gemini-1.5-Pro as the judge model for evaluating reasoning quality, for more detailed evaluation criteria and prompt, please refer to Appendix B.

\begin{figure}[!t]
    \centering
    \includegraphics[width=\textwidth]{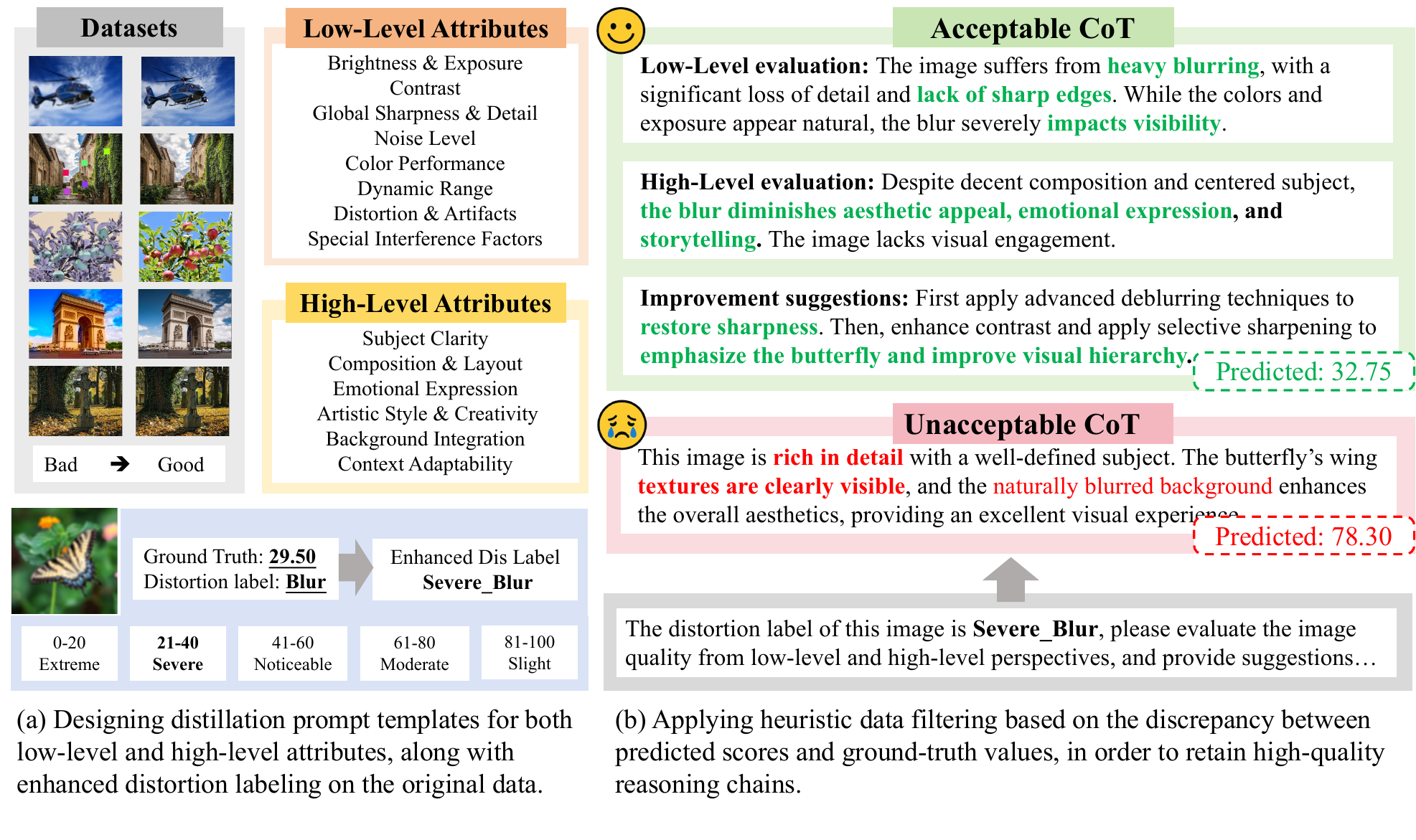}
    \caption{\textbf{Illustration of the distillation process.} 
    (a) We design distillation prompts targeting both low-level (e.g., sharpness, color, artifacts) and high-level (e.g., composition, emotional expression) attributes, with enhanced distortion labeling (e.g., Severe\_Blur) added to the dataset. 
    (b) We filter generated reasoning chains based on the alignment between predicted and ground-truth scores.}
    \label{fig2}
    \vspace{-5mm}
\end{figure}

\vspace{-1mm}
\subsection{Cold start}
\label{Cold start}
In general-purpose MLLMs that have not undergone supervised fine-tuning (SFT), models with smaller parameter scales often struggle to match the fine-grained perception and descriptive capabilities of larger models. To ensure that Qwen2.5VL-7B possesses sufficient reasoning ability before reinforcement fine-tuning—and to reduce early-stage reasoning confusion and severe hallucination during the reinforcement process—we introduce a cold-start initialization step to warm up our baseline model. In this process, we first define an expert-designed prompt template (Fig.\ref{fig2}(a), prompting the teacher model to assess image quality from both low-level and high-level perspectives for a curated set of images and output raw quality scores. Simultaneously, if the dataset contains predefined distortion type labels, we enhance them with corresponding descriptive adjectives and feed them as additional hints to the teacher model. To ensure the rationality and reliability of the data used for reasoning distillation, we filter out samples that contain clear logical flaws in their reasoning chains (Fig.\ref{fig2}(b)) using a heuristic selection rule:
\begin{equation}
\mathcal{D}_{\text{refined}}
= \left\{ i \,\middle|\, \text{rank}\left(|\hat{y}_i - y_i|\right) \leq \gamma \cdot N \right\},
\end{equation}
where $\hat{y}_i$ and $y_i$ denote the predicted score and ground truth for the $i$-th sample; $\text{rank}(\cdot)$ ranks the absolute error values in ascending order; $\gamma \in (0, 1)$ represents the percentage of samples to retain (e.g., $\gamma=0.8$ means keeping the top 80\% with smallest errors); and $N$ is the total number of samples.

Using this filtering rule, we observe a significant improvement in the quality and reliability of the data used for reasoning distillation. For example, by filtering only 20\% of the most inconsistent samples, the overall PLCC/SRCC of the distilled data improves from 0.807 / 0.561 to 0.922 / 0.795, and the reasoning quality scores (completeness, accuracy, reasonableness) increase from 4.73 / 3.41 / 4.20 to 4.75 / 3.61 / 4.45 (Sec.~\ref{Preliminaries}).By replacing the final score labels with the ground truth, we perform SFT on Qwen2.5VL-7B using this curated dataset. The resulting model is referred to as \textbf{Q-Ponder-CI (Cold-start Initialization)}.The full prompt templates used for reasoning distillation are provided in the Appendix A.

\begin{figure}[htbp]
    \centering
    \includegraphics[width=\textwidth]{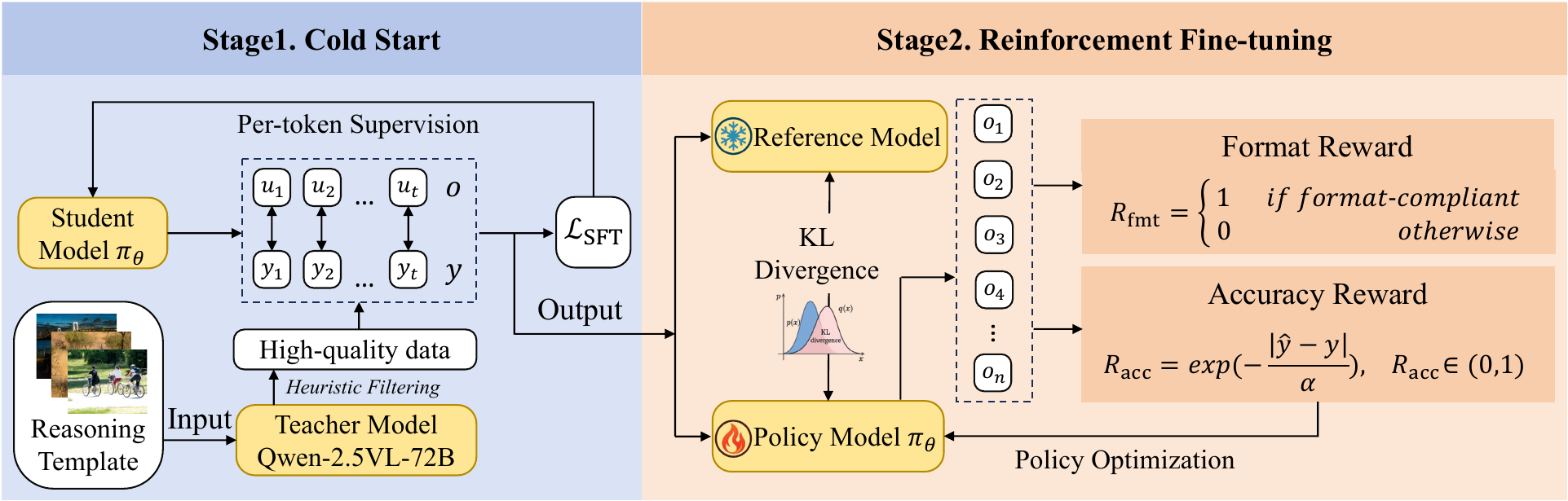}
    \caption{The policy model is first initialized via cold start to enhance reasoning. It then generates multiple trajectories, which are evaluated by a reward model and used for joint policy optimization.}
    \label{fig3}
    \vspace{-10pt}
\end{figure}

\subsection{Reasoning-oriented reinforcement learning}
\label{Reasoning-oriented reinforcement learning}

After cold-start initialization, \textbf{Q-Ponder-CI} demonstrates basic reasoning and score prediction capabilities (Sec.~\ref{Cold start}). However, hallucinations and incorrect reasoning about distortions still occur (see Appendix D), likely due to the limitations of SFT, which captures linguistic patterns without directly supervising numerical outputs. Additionally, there remains frequent confusion regarding the output format. To address these issues, we introduce a second-stage reinforcement learning phase based on the GRPO algorithm to jointly supervise reasoning format and score accuracy (Fig.~\ref{fig3}).

Compared to the commonly used PPO algorithm, GRPO does not rely on the traditional critic network (value function estimator). Instead, it estimates the baseline directly from a set of sample scores by computing advantages based on relative scores of a group of actions to guide policy optimization. Specifically, for each query $q$, GRPO samples a set of outputs $\{o_1, o_2, \cdots, o_n\}$ from the old policy $\pi_{\theta_{old}}$ and optimizes the new policy $\pi_\theta$ by maximizing the following objective function:

\vspace{-0.5cm}
\begin{align}
\label{eq2}
&\mathcal{I}_{GRPO}(\theta) = \operatorname{\mathbb{E}}_{(I,\mathbf{q}) \sim p_{\mathcal{D}},\, \{\mathbf{o}_i\}_{i=1}^n \sim \pi_{\theta_{\text{old}}}(\cdot|I,\mathbf{q})}
\bigg[
\frac{1}{n} \sum_{i=1}^n \bigg(
\min\bigg(
\frac{\pi_\theta(\mathbf{o}_i|I,\mathbf{q})}{\pi_{\theta_{\text{old}}}(\mathbf{o}_i|I,\mathbf{q})} \hat{A}_i, \notag \\
&\quad \operatorname{clip}\left(
\frac{\pi_\theta(\mathbf{o}_i|I,\mathbf{q})}{\pi_{\theta_{\text{old}}}(\mathbf{o}_i|I,\mathbf{q})},
1{-}\varepsilon, 1{+}\varepsilon
\right) \hat{A}_i
\bigg)
- \beta\, \mathbb{D}_{KL}(\pi_\theta(\cdot|I,\mathbf{q}) \,\|\, \pi_{\text{ref}}(\cdot|I,\mathbf{q}))
\bigg)
\bigg],
\end{align}
\vspace{-3mm}

where $\varepsilon$ and $\beta$ are hyperparameters. The advantage $\hat{A}_i$ is computed via z-score normalization of rewards within each sampled output group:

\vspace{-2mm}
\begin{equation}
\label{eq:advantage}
\hat{A}_i = \frac{r_i - \max(\{r_1, r_2, \cdots, r_n\})}{\operatorname{std}(\{r_1, r_2, \cdots, r_n\})},
\end{equation}

which provides a relative advantage estimate, encouraging the model to prefer outputs that are better than their group-wise peers rather than relying on absolute rewards.

To compute the policy update, we first evaluate the importance ratio $\textit{ratio}^{(i)} = \frac{\pi_\theta(\mathbf{o}_i|I,\mathbf{q})}{\pi_{\theta_{\text{old}}}(\mathbf{o}_i|I,\mathbf{q})}$ for each sampled output. To stabilize training and avoid overly large policy updates, $\textit{ratio}^{(i)}$ is clipped within $[1-\varepsilon, 1+\varepsilon]$. The clipped ratio is then multiplied with the estimated advantage to compute the surrogate loss. In addition, we regularize the new policy to stay close to a reference policy $\pi_{\text{ref}}$ (often chosen as $\pi_{\theta_{\text{old}}}$ or a separately trained baseline) via a KL divergence penalty weighted by $\beta$:

\vspace{-3mm}
\begin{equation}
\label{eq:kl}
\mathbb{D}_{\mathrm{KL}}(\pi_\theta || \pi_{\text{ref}}) = \frac{\pi_{\text{ref}}(\mathbf{o}_i|I,\mathbf{q})}{\pi_\theta(\mathbf{o}_i|I,\mathbf{q})}
- \log\frac{\pi_{\text{ref}}(\mathbf{o}_i|I,\mathbf{q})}{\pi_\theta(\mathbf{o}_i|I,\mathbf{q})} - 1,
\end{equation}

\vspace{-1mm}
The reward function plays a critical role in guiding GRPO training by producing the reward signals $\{r_1, r_2, \cdots, r_n\}$ for each group of outputs during rollout. In classical reasoning tasks, rewards are often sparse and binary, typically determined by whether the final answer is correct. However, in IQA, quality labels (e.g., MOS or DMOS) are continuous and subjective, making such sparse reward signals inadequate for guiding learning effectively—especially when the model’s initial reasoning ability is limited. Sparse rewards provide little gradient signal for sub-optimal predictions, which may cause the model to frequently receive zero feedback during rollouts and get trapped in a local optimum with no learning progress. To mitigate the drawbacks of sparse reward signals, we introduce a dense and smooth accuracy reward. This reward encourages gradual improvements by softly measuring the discrepancy between the predicted score $\hat{y}$ and the ground-truth score $y$ as follows:

\vspace{-4mm}
\begin{equation}
\label{eq5}
R_{\text{acc}} = \exp\left(-\frac{|\hat{y}-y|}{\alpha}\right),\quad R_{\text{acc}} \in (0, 1),
\end{equation}

\vspace{-2mm}
where $\alpha$ is a tolerance hyperparameter, with smaller values imposing sharper penalties for deviations.

\vspace{-2mm}
Moreover, to address frequent format errors in reasoning chains (e.g., missing tags, malformed steps) observed in Q-Ponder-CI, we introduce a format compliance reward to enforce output structure. The format reward checks whether reasoning content is enclosed within ``\texttt{<think>}...<\texttt{/think>}'' tags and whether final answers are correctly wrapped in ``\texttt{<answer>}...<\texttt{/answer>}'':

\vspace{-5mm}
\begin{equation}
\label{eq:fmt-reward}
R_{\text{fmt}} =
\begin{cases}
1 & \text{if format-compliant} \\
0 & \text{otherwise}
\end{cases}
\end{equation}

\vspace{-2mm}
Finally, we combine these reward types into a weighted total reward:

\vspace{-4mm}
\begin{equation}
\label{eq7}
R_{\text{total}} = \lambda_{\text{acc}} \cdot R_{\text{acc}} + \lambda_{\text{fmt}} \cdot R_{\text{fmt}},
\end{equation}

\vspace{-2mm}
where coefficients $\lambda_{\text{acc}}$ and $\lambda_{\text{fmt}}$ weight score accuracy and format compliance in the final reward.

\section{Experiments}
\subsection{Experiment setups}
\label{sec4.1}
\paragraph{IQA Datasets.} For the score regression task, we conduct training and evaluation on six IQA datasets grouped into three categories:  (1) \textit{In-the-wild datasets}, including KonIQ~\cite{hosu2020koniq}, SPAQ~\cite{fang2020perceptual}, and LIVE-Wild~\cite{ghadiyaram2015live};  (2) \textit{Synthetic distortion datasets}, including KADID~\cite{lin2019kadid} and CSIQ~\cite{larson2010most};  (3) \textit{AI-generated image datasets}, including AGIQA~\cite{li2023agiqa}. We follow the same split as in~\cite{you2025teaching} for the KonIQ dataset, using 80\% of the filtered training set (approximately 5,600 images) for training.
For the evaluation of reasoning quality in chain-of-thought descriptions, we conduct experiments on six datasets: KADID~\cite{lin2019kadid}, CIDIQ~\cite{liu2014cid}, CSIQ~\cite{larson2010most}, LIVE~\cite{sheikh2006statistical}, TID2008~\cite{ponomarenko2009tid2008}, and TID2013~\cite{ponomarenko2015image}. All datasets used for reasoning evaluation are distinct from the training set and thus considered OOD. Notably, these six datasets are all derived from synthetic distortion datasets with a rich variety of predefined distortion types, making them well-suited for evaluating the model’s reasoning ability under diverse degradation conditions. For both quality score regression and reasoning description evaluation, we follow the evaluation metrics detailed in Sec. \ref{Preliminaries}.

\paragraph{Implementation Details.} 
We initialize Qwen2.5-VL-7B and conduct SFT using cross-entropy loss. The model with the best OOD regression performance is selected as Q-Ponder-CI, which typically converges within 300 steps. In the second-stage RL, we set the sampling number $n$ in Eqn.~(\ref{eq2}) to 8, with hyperparameters $\beta = 0.04$ and $\epsilon = 0.2$. The reward decay factor in Eqn.~(\ref{eq5}) is set to $\alpha = 10$, and the reward weights in Eqn.~(\ref{eq7}) are $\lambda_{\text{acc}} = 2$ and $\lambda_{\text{fmt}} = 1$. We use a batch size of 1 and an initial learning rate of $2 \times 10^{-5}$ with cosine decay scheduling. The entire RL stage converges within one epoch and completes in three days using 8 NVIDIA A100 GPUs.

\subsection{Main results}

\begin{figure}[htbp]
    \centering
    \includegraphics[width=\textwidth]{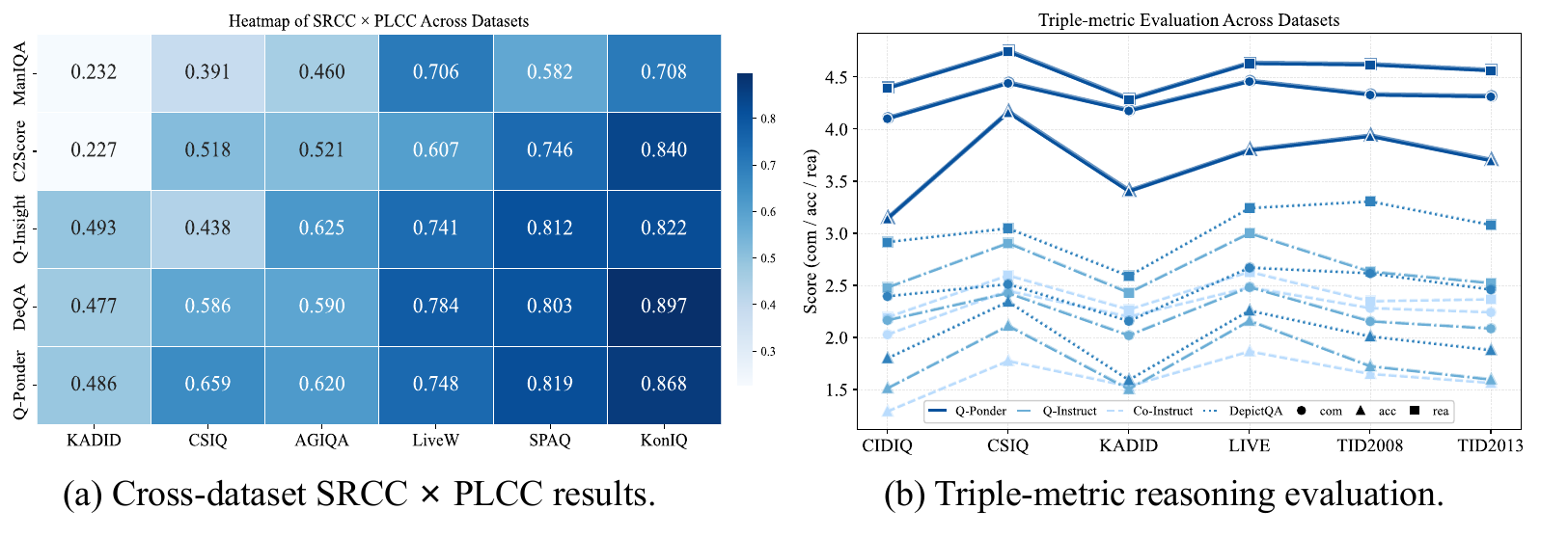}
    \caption{\textbf{Illustration of main evaluation results.} (a) Heatmap showing cross-dataset performance on quality score prediction (SRCC × PLCC). (b) Line plot comparing the completeness, accuracy, and reasonableness of reasoning descriptions across datasets for different models.
}
    \label{fig4}
    \vspace{-10pt}
\end{figure}

\paragraph{Image Quality Score Regression.} Tab.~\ref{tab1} and Fig.~\ref{fig4}(a) compare the performance of our model, Q-Ponder, with existing SOTA score regression methods across three categories: \textit{handcrafted}, \textit{deep learning-based}, and \textit{MLLM-based} models (for Q-Insight, we report the scores-only version trained on the KonIQ dataset). All models are matched in parameter size and trained on the same 7k KonIQ subset for fairness. Q-Ponder consistently outperforms other models on both in-domain and OOD datasets. In particular, we observe significant performance gains on two OOD datasets—CSIQ and AGIQA, which we attribute to two factors:  
\textbf{(i)} Q-Ponder integrates high-quality reasoning chains during cold-start, along with a dense reward function that stabilizes gradients and enhances training feedback;  
\textbf{(ii)} In the reinforcement fine-tuning stage, the model learns from negative samples and generalizes scoring as a policy, rather than merely fitting input-output pairs via token-level cross-entropy loss (Sec.~\ref{Reasoning-oriented reinforcement learning}).

The only exception occurs on the in-domain KonIQ dataset, where our model slightly underperforms pure score regression methods like Q-Align and DeQA by a margin of ~0.01. This reflects a deliberate trade-off in our training strategy: during the selection of the cold-start model, we observed that continued training could further improve in-domain performance, but at the cost of significant degradation on all OOD datasets, along with increased hallucinations and instability in the generated reasoning chains. To ensure robustness and generalizability under limited supervision, we prioritized OOD performance and reasoning consistency when selecting the pre-warmed model. Although this choice slightly compromises KonIQ accuracy, it yields stronger reliability across diverse real-world scenarios, which we believe is essential for practical IQA applications.

\vspace{-3mm}

\begin{table}[htbp]
\centering
\caption{
Performance comparison of IQA methods across different dataset categories. Each cell shows PLCC / SRCC. \textbf{Top-1} and \textbf{Top-2} results are marked with 
\colorbox{LightRed}{\textbf{red}}/\colorbox{LightBlue}{\textbf{blue}} background colors.
}

\label{tab1}

\renewcommand{\arraystretch}{1.1}

\resizebox{\textwidth}{!}{%
\begin{tabular}{lccccccc}
\toprule
\multirow{2}{*}{\centering\arraybackslash\raisebox{-1ex}{\textbf{Methods}}} & 
\multicolumn{3}{c}{\textbf{Wild Images}} & 
\multicolumn{2}{c}{\textbf{Synthetic Distortion}} & 
\multicolumn{1}{c}{\textbf{AI-Generated}} & 
\multirow{2}{*}{\centering\arraybackslash\raisebox{-1ex}{\textbf{AVG.}}}
 \\
\cmidrule(lr){2-4} \cmidrule(lr){5-6} \cmidrule(lr){7-7}
& KonIQ\cite{hosu2020koniq} & SPAQ\cite{fang2020perceptual} & LiveW\cite{ghadiyaram2015live} & KADID\cite{lin2019kadid} & CSIQ\cite{larson2010most} & AGIQA\cite{li2023agiqa} & \\

\midrule
\multicolumn{8}{c}{\textbf{Handcrafted Methods}} \\
\midrule
NIQE~\cite{mittal2012making}    & 0.533 / 0.530 & 0.679 / 0.664 & 0.493 / 0.449 & 0.468 / 0.405 & 0.718 / 0.628 & 0.560 / 0.533 & 0.575 / 0.535 \\
BRISQUE~\cite{mittal2012no}     & 0.225 / 0.226 & 0.490 / 0.406 & 0.361 / 0.313 & 0.429 / 0.356 & 0.740 / 0.556 & 0.541 / 0.497 & 0.464 / 0.392 \\

\midrule
\multicolumn{8}{c}{\textbf{Deep-learning Methods}} \\
\midrule
NIMA~\cite{talebi2018nima}         & 0.896 / 0.859 & 0.838 / 0.856 & 0.814 / 0.711 & 0.532 / 0.535 & 0.695 / 0.649 & 0.715 / 0.654 & 0.748 / 0.711 \\
HyperIQA~\cite{su2020blindly}     & 0.917 / 0.906 & 0.791 / 0.788 & 0.772 / 0.701 & 0.506 / 0.468 & 0.752 / 0.717 & 0.702 / 0.640 & 0.740 / 0.703 \\
DBCNN~\cite{networkblind}         & 0.884 / 0.875 & 0.812 / 0.806 & 0.773 / 0.755 & 0.497 / 0.484 & 0.586 / 0.572 & 0.730 / 0.641 & 0.714 / 0.689 \\
MUSIQ~\cite{ke2021musiq}        & 0.924 / 0.929 & 0.868 / 0.863 & 0.789 / 0.830 & 0.575 / 0.556 & 0.771 / 0.710 & 0.722 / 0.630 & 0.775 / 0.753 \\
CLIP-IQA+~\cite{wang2023exploring} & 0.909 / 0.895 & 0.866 / 0.854 & 0.832 / 0.805 & 0.653 / 0.642 & 0.772 / 0.719 & 0.736 / 0.685 & 0.795 / 0.767 \\
ManIQA~\cite{yang2022maniqa}      & 0.849 / 0.834 & 0.768 / 0.758 & 0.849 / 0.832 & 0.499 / 0.465 & 0.623 / 0.627 & 0.723 / 0.636 & 0.719 / 0.692 \\

\midrule
\multicolumn{8}{c}{\textbf{MLLM-based Methods}} \\
\midrule
C2Score~\cite{zhu2024adaptive}     & 0.923 / 0.910 & 0.867 / 0.860 & 0.786 / 0.772 & 0.500 / 0.453 & 0.735 / 0.705 & 0.777 / 0.671 & 0.765 / 0.729 \\
Q-Align~\cite{wu2023q}         & \cellcolor{LightBlue}0.941 / 0.940 & 0.886 / 0.887 & 0.853 / 0.860 & 0.674 / 0.684 & 0.671 / 0.737 & 0.772 / 0.735 & 0.799 / 0.807 \\
DeQA~\cite{you2025teaching}      & \cellcolor{LightRed}0.953 / 0.941 & 0.895 / 0.896 & \cellcolor{LightRed}0.892 / 0.879 & 0.694 / 0.687 & \cellcolor{LightBlue}0.787 / 0.744 & 0.809 / 0.729 & \cellcolor{LightBlue}0.838 / 0.813 \\
Q-Insight~\cite{li2025q}           & 0.918 / 0.895 & \cellcolor{LightBlue}0.903 / 0.899 & 0.870 / 0.839 & \cellcolor{LightRed}0.702 / 0.702 & 0.685 / 0.640 & \cellcolor{LightBlue}0.816 / 0.766 & 0.816 / 0.790 \\
\textbf{Q-Ponder}                  & \textbf{0.937 / 0.926} & \cellcolor{LightRed}\textbf{0.904 / 0.906} & \cellcolor{LightBlue}\textbf{0.882 / 0.848} & \cellcolor{LightBlue}\textbf{0.693 / 0.701} & \cellcolor{LightRed}\textbf{0.832 / 0.792} & \cellcolor{LightRed}\textbf{0.821 / 0.755} & \cellcolor{LightRed}\textbf{0.845 / 0.821} \\

\bottomrule
\end{tabular}%
}
\vspace{-1mm}
\end{table}

\textbf{Image Quality Description.}
Tab.\ref{tab2} further validates the effectiveness of the reasoning chains generated by Q-Ponder, evaluated under the image quality description task. We compare Q-Ponder against several SOTA description-based IQA models. (Q-Insight is excluded since it has not been open-sourced.) To ensure fairness, all models are guided using the same prompt templates employed during the distillation stage. As shown in Fig.\ref{fig4}(b), Q-Ponder outperforms all baselines, achieving the highest completeness, accuracy, and reasonableness across all datasets.
By comparing the reasoning outputs (see Appendix D), we observe that prior description-based models tend to focus on specific tasks (Sec.~\ref{evalR}), which limits their reasoning capabilities to certain distortion types or simple global quality judgments. This reduces their ability to handle complex distortion scenarios and to provide deep, nuanced insights. Such behavior can be interpreted as a form of reasoning-level overfitting.
\vspace{-1mm}

\definecolor{LightGray}{gray}{0.93}
\vspace{-3mm}
\begin{table}[htbp]
\centering
\captionsetup{justification=raggedright,
              singlelinecheck=false,
              width=\textwidth}
\caption{Quantitative comparison of reasoning descriptions across six datasets (com / acc / rea).}
\small
\renewcommand{\arraystretch}{1.2}
\setlength{\tabcolsep}{3pt}

\resizebox{\textwidth}{!}{%
\begin{tabular}{@{}lccccccc@{}}
\toprule
Methods & CIDIQ\cite{liu2014cid} & CSIQ\cite{larson2010most} & KADID\cite{lin2019kadid} & LIVE\cite{sheikh2006statistical} & TID2008\cite{ponomarenko2009tid2008} & TID2013\cite{ponomarenko2015image} & AVG. \\
\midrule
Co-Instruct\cite{wu2024towards}
  & 2.026 / 1.288 / 2.192 & 2.446 / 1.772 / 2.594 & 2.198 / 1.530 / 2.266 & 2.479 / 1.864 / 2.628 & 2.280 / 1.648 / 2.346 & 2.240 / 1.562 / 2.364 & 2.096 \\
Q-Instruct\cite{wu2024q}
  & 2.164 / 1.510 / 2.478 & 2.424 / 2.110 / 2.904 & 2.018 / 1.502 / 2.428 & 2.481 / 2.158 / 2.999 & 2.153 / 1.722 / 2.630 & 2.084 / 1.594 / 2.520 & 2.216 \\
DepictQA\cite{you2024depicting}
  & 2.394 / 1.798 / 2.914 & 2.510 / 2.342 / 3.046 & 2.156 / 1.590 / 2.588 & 2.668 / 2.256 / 3.240 & 2.614 / 2.008 / 3.304 & 2.458 / 1.876 / 3.078 & 2.454 \\
\textbf{Q-Ponder} 
  & \textbf{4.100 / 3.140 / 4.392} & \textbf{4.440 / 4.160 / 4.744} & \textbf{4.175 / 3.402 / 4.283} & \textbf{4.457 / 3.794 / 4.630} & \textbf{4.328 / 3.931 / 4.617} & \textbf{4.311 / 3.697 / 4.561} & \textbf{4.182} \\
\bottomrule
\end{tabular}
}
\label{tab2}
\vspace{-3mm}
\end{table}

\subsection{Ablation studies}
\paragraph{Effectiveness of Reasoning Chains and Reinforcement Fine-Tuning on score regression.} To assess the impact of reasoning chains and RL on score regression, we conducted ablation experiments under four configurations: (1) SFT without reasoning chains, (2) SFT with short reasoning chains (100--200 tokens), (3) SFT with expert-level CoT reasoning (700--800 tokens, denoted as Q-Ponder-CI), and (4) RL fine-tuning based on Q-Ponder-CI (i.e., Q-Ponder). All other variables were kept constant. As shown in Tab.~\ref{tab3}, longer and more comprehensive reasoning chains significantly improve performance, especially on OOD datasets like AGIQA and LIVEW. RL further enhances the model’s regression ability beyond the strong cold-start baseline.
\vspace{-10pt}

\newcommand{\cmark}{\ding{51}}  
\newcommand{\xmark}{\ding{55}}  

\begin{table}[htbp]
\centering
\caption{Impact of CoT Length and Reinforcement Learning on Score Regression (PLCC / SRCC).}
\small
\renewcommand{\arraystretch}{1.2}
\setlength{\tabcolsep}{3pt}
\resizebox{\linewidth}{!}{
\begin{tabular}{ccc|ccccccc}
\toprule
\textbf{CoT} & \textbf{SFT} & \textbf{RL} & KonIQ\cite{hosu2020koniq} & SPAQ\cite{fang2020perceptual} & LiveW\cite{ghadiyaram2015live} & KADID\cite{lin2019kadid} & CSIQ\cite{larson2010most} & AGIQA\cite{li2023agiqa} & AVG. \\
\midrule
\xmark & \cmark & \xmark & 0.889 / 0.866 & 0.874 / 0.875 & 0.734 / 0.728 & 0.668 / 0.663 & 0.813 / 0.739 & 0.674 / 0.650 & 0.775 / 0.754 \\
short  & \cmark & \xmark & 0.894 / 0.905 & 0.873 / 0.883 & 0.813 / 0.801 & 0.665 / 0.648 & 0.721 / 0.700 & 0.749 / 0.707 & 0.786 / 0.774 \\
long   & \cmark & \xmark & 0.917 / 0.902 & 0.908 / 0.902 & 0.851 / 0.819 & 0.688 / 0.661 & 0.822 / 0.775 & 0.816 / 0.742 & 0.834 / 0.800 \\
long   & \cmark & \cmark & \textbf{0.937 / 0.926} & \textbf{0.904 / 0.906} & \textbf{0.882 / 0.848} & \textbf{0.693 / 0.701} & \textbf{0.832 / 0.792} & \textbf{0.821 / 0.755} & \textbf{0.845 / 0.821} \\
\bottomrule
\end{tabular}
}
\label{tab3}
\end{table}

\vspace{-3mm}
\paragraph{Impact of Two-Stage Training Pipeline on Reasoning Descriptions.} To further demonstrate that our training pipeline enhances the model’s reasoning descriptions, we compare the performance of (1) baseline model (Qwen2.5VL-7B), (2) teacher model (Qwen2.5VL-72B), (3) SFT model with expert reasoning (Q-Ponder-CI), and (4) final model with RL fine-tuning (Q-Ponder) on the reasoning description evaluation benchmark. As shown in Tab.~\ref{tab4}, we observe that models trained with both stages achieve significant improvements over the baseline. Q-Ponder outperforms both Q-Ponder-CI and the teacher model in terms of accuracy and reasonableness, though with a slight drop in completeness. We attribute this to the fact that both the teacher model and the distilled Q-Ponder-CI tend to generate more elaborate and exhaustive reasoning chains to cover all dimensions, which, however, increases the risk of hallucinations and inaccurate judgments. In contrast, RL optimizes the reasoning logic by slightly reducing unnecessary reasoning content, thereby improving the reasonableness and accuracy of the analysis. Appendix C presents the variation in reasoning lengths during training, while the case analyses in Appendix D further support these observations.

\vspace{-3mm}

\begin{table}[htbp]
\centering
\caption{Evaluation of Reasoning Descriptions via Triple-Metric Analysis (Com / Acc / Rea)}
\small
\renewcommand{\arraystretch}{1.2}
\setlength{\tabcolsep}{4pt}
\resizebox{\linewidth}{!}{
\begin{tabular}{ccc|ccccccc}
\toprule
Param & SFT & RL & CIDIQ\cite{liu2014cid} & CSIQ\cite{larson2010most} & KADID\cite{lin2019kadid} & LIVE\cite{sheikh2006statistical} & TID2008\cite{ponomarenko2009tid2008} & TID2013\cite{ponomarenko2015image} & AVG. \\
\midrule
7B  & \xmark & \xmark & 4.094 / 2.000 / 3.952 & 4.414 / 3.096 / 4.298 & 4.098 / 2.398 / 3.908 & 4.298 / 2.522 / 4.119 & 4.242 / 3.087 / 4.206 & 4.195 / 2.829 / 4.217 & 3.665 \\
72B & \xmark & \xmark & 4.321 / 2.642 / 4.101 & 4.725 / 3.408 / 4.195 & 4.573 / 3.173 / 4.154 & 4.573 / 3.173 / 4.154 & 4.539 / 3.425 / 4.122 & 4.540 / 3.345 / 4.178 & 3.963 \\
7B  & \cmark & \xmark & 4.408 / 2.654 / 4.308 & 4.754 / 3.432 / 4.514 & 4.373 / 2.862 / 4.118 & 4.654 / 3.205 / 4.362 & 4.567 / 3.202 / 4.357 & 4.526 / 3.030 / 4.369 & 3.983 \\
7B  & \cmark & \cmark & \textbf{4.100 / 3.140 / 4.392} & \textbf{4.440 / 4.160 / 4.744} & \textbf{4.175 / 3.402 / 4.283} & \textbf{4.457 / 3.794 / 4.630} & \textbf{4.328 / 3.931 / 4.617} & \textbf{4.311 / 3.697 / 4.561} & \textbf{4.182} \\
\bottomrule
\end{tabular}
}
\label{tab4}
\vspace{-4mm}
\end{table}

\subsection{Conclusion and Limitation}
\paragraph{Conclusion.} In summary, we have introduced a unified cold-start + RL training pipeline based on GRPO, establishing a new paradigm for training image quality assessment models. Our approach enables the model to simultaneously develop two core capabilities: accurate quality score regression and coherent, reasonable reasoning, even with limited annotated data. The resulting model, Q-Ponder, demonstrates strong performance in both numerical accuracy and descriptive reasoning across a wide range of datasets, significantly outperforming prior SOTA methods, especially under OOD scenarios. Looking ahead, we aim to extend this training framework to other vision-language tasks such as aesthetic quality assessment and temporally-aware video quality evaluation, providing stronger baselines and insights for future image restoration and autonomous visual agents.

\vspace{-3mm}
\paragraph{Limitation.}
The quality of reasoning descriptions is currently judged by a single LLM, without secondary verification from human experts or other proprietary
 MLLMs (e.g., OpenAI o3). This lowers confidence in the results and leaves potential model biases unaddressed. We plan to introduce multi-source reviewing and bias-detection mechanisms to improve evaluation reliability.

\small
\bibliography{reference}

\newpage
\appendix

\section{Prompt Templates for Reasoning Distillation}
\label{A}
\paragraph{Prompt Structure Overview.}
The table below summarizes the structured prompt used during the reasoning distillation stage. It guides the teacher MLLM to analyze both low-level distortions and high-level aesthetics in a step-by-step manner, optionally incorporating hidden distortion hints.

\begin{table}[h]
\centering
\caption{Prompt Structure for Reasoning Distillation}
\renewcommand{\arraystretch}{1.3}
\small
\begin{tabular}{p{0.20\linewidth} | p{0.75\linewidth}}
\toprule
\textbf{Section} & \textbf{Prompt Content} \\
\midrule
\textbf{System Prompt} &
You are an expert in image quality and aesthetic evaluation. Your task is to comprehensively analyze and evaluate image quality based on low-level physical attributes and high-level aesthetic attributes through detailed, step-by-step chain-of-thought reasoning. Please strictly follow the judgment criteria below to gradually expand your reasoning and discussion. \\
\midrule
\textbf{Low-Level Attribute Analysis} & 
\textit{Note: These are critical technical factors and must be fully addressed.} \newline
1. Brightness and Exposure \newline
2. Contrast Evaluation \newline
3. Sharpness and Detail Preservation \newline
4. Noise Level (digital vs. compression noise) \newline
5. Color Performance (WB, saturation, transitions) \newline
6. Dynamic Range \newline
7. Distortion and Artifacts (e.g., lens, moiré) \newline
8. Visual Interference (e.g., glare, weather, clutter) \\
\midrule
\textbf{High-Level Attribute Analysis} & 
\textit{Note: These are supportive attributes, subordinate to low-level analysis.} \newline
Special rule: Artistic style may conditionally override low-level flaws, but must not affect final score. \newline
1. Subject Clarity (focus, separation) \newline
2. Composition and Layout (symmetry, rule of thirds, balance) \newline
3. Emotional Expression and Storytelling \newline
4. Artistic Style and Creativity \newline
5. Environment and Background Integration \newline
6. Context Adaptability (e.g., portrait vs. landscape) \\
\midrule
\textbf{Comprehensive Evaluation and Output Format} & 
Summarize the analysis and provide concrete improvement suggestions. \newline
Clearly explain the reasoning basis and logic. \newline
Structure the output into three numbered sections: \newline
(1) Low-level analysis + distortion severity \newline
(2) High-level analysis \newline
(3) Final suggestions \newline
Wrap everything in \texttt{<think>...</think>}, followed by a final score in \texttt{<answer>...</answer>} between 0 and 100. \\
\midrule
\textbf{Distortion Tag Hint (Optional)} &
I will give you a hint: the soft label for this image is \textcolor{red}{[ distortion\_tags ]}, which may suggest its potential distortion type and severity. You should carefully take this into account during the low-level attribute analysis.  \\
\bottomrule
\end{tabular}
\end{table}

\paragraph{Attribution of Evaluation Criteria.}
The evaluation indicators used in our reasoning prompt are grounded in established literature from both IQA and aesthetic evaluation. 

For \textbf{low-level attributes}, such as brightness, contrast, sharpness, noise, dynamic range, color fidelity, and artifacts, our design follows conventional IQA standards and visual perception literature. Key references include: 
\begin{itemize}
    \item \textit{Imatest Image Quality Factors}, which outlines physical quality metrics like sharpness, dynamic range, noise, distortion, and color accuracy.\footnote{\url{https://www.imatest.com/docs/iqfactors/}}
    \item The Wikipedia entry on \textit{Image Quality}, which summarizes perceptual attributes used in traditional and modern IQA systems.\footnote{\url{https://en.wikipedia.org/wiki/Image_quality}}
\end{itemize}

For \textbf{high-level aesthetic attributes}, such as composition, subject clarity, storytelling, emotional tone, and artistic style, we draw from the image aesthetics and computational photography literature. Notable sources include:
\begin{itemize}
    \item \textit{Composition and Style Attributes Guided Image Aesthetic Assessment}, which formalizes semantic and stylistic attributes for aesthetic quality modeling.\footnote{\url{https://arxiv.org/abs/2111.04647}}
    \item \textit{Understanding the Importance of Artistic Aspects of Camera Image Quality}, which emphasizes creativity and artistic criteria in professional image evaluation.\footnote{\url{https://library.imaging.org/ei/33/9/art00002}}
\end{itemize}

These sources collectively form the theoretical foundation for the step-by-step reasoning structure used during the distillation process in our training pipeline.

\section{Prompt Templates for Reasoning Evaluation}
\label{B}
\paragraph{Evaluation Prompt for Reasoning Quality Assessment.}
We design the following template to assess the quality of the student model’s reasoning chain. The evaluator must score completeness, accuracy, and reasonableness based on the image’s distortion tag and the generated explanation.
\begin{table}[h]
\centering
\caption{Prompt Template for Reasoning Evaluation}
\renewcommand{\arraystretch}{1.3}
\small
\begin{tabular}{p{0.25\linewidth} | p{0.70\linewidth}}
\toprule
\textbf{Section} & \textbf{Prompt Content} \\
\midrule
\textbf{System Instruction} &
You are an expert image-quality evaluator. Your task is to evaluate the quality of a model's reasoning response based on the image and its distortion tag. \\
\midrule
\textbf{Model Response} &
\verb|<answer>{student_answer}</answer>| \\
\midrule
\textbf{Distortion Tag Input} &
\verb|<tag>{distortion_tag}</tag>| \\
\midrule
\textbf{Task 1: Completeness} &
Assess whether the response covers all relevant distortion aspects in the image and demonstrates multiple evaluation perspectives. Rate on a scale of [1–5]. \\
\midrule
\textbf{Task 2: Accuracy} &
Evaluate whether the response correctly identifies key distortions—especially those relevant to the distortion tag—and whether the severity level is appropriate. Rate on a scale of [1–5]. \\
\midrule
\textbf{Task 3: Reasonableness} &
Check the logical consistency of the reasoning, and ensure it is free from contradictions, hallucinations, or misjudgments. Rate on a scale of [1–5]. \\
\midrule
\textbf{Output Format} &
The final output must follow this exact XML-style format: \newline
\verb|<Completeness>score</Completeness>| \newline
\verb|<Accuracy>score</Accuracy>| \newline
\verb|<Reasonableness>score</Reasonableness>| \newline
\textit{Example:} \newline
\verb|<Completeness>3</Completeness>| \newline
\verb|<Accuracy>4</Accuracy>| \newline
\verb|<Reasonableness>5</Reasonableness>| \\
\bottomrule
\end{tabular}
\end{table}

\section{Training Dynamics and Reasoning Evolution}
\label{C}

\begin{figure}[h]
    \centering
    \includegraphics[width=0.95\textwidth]{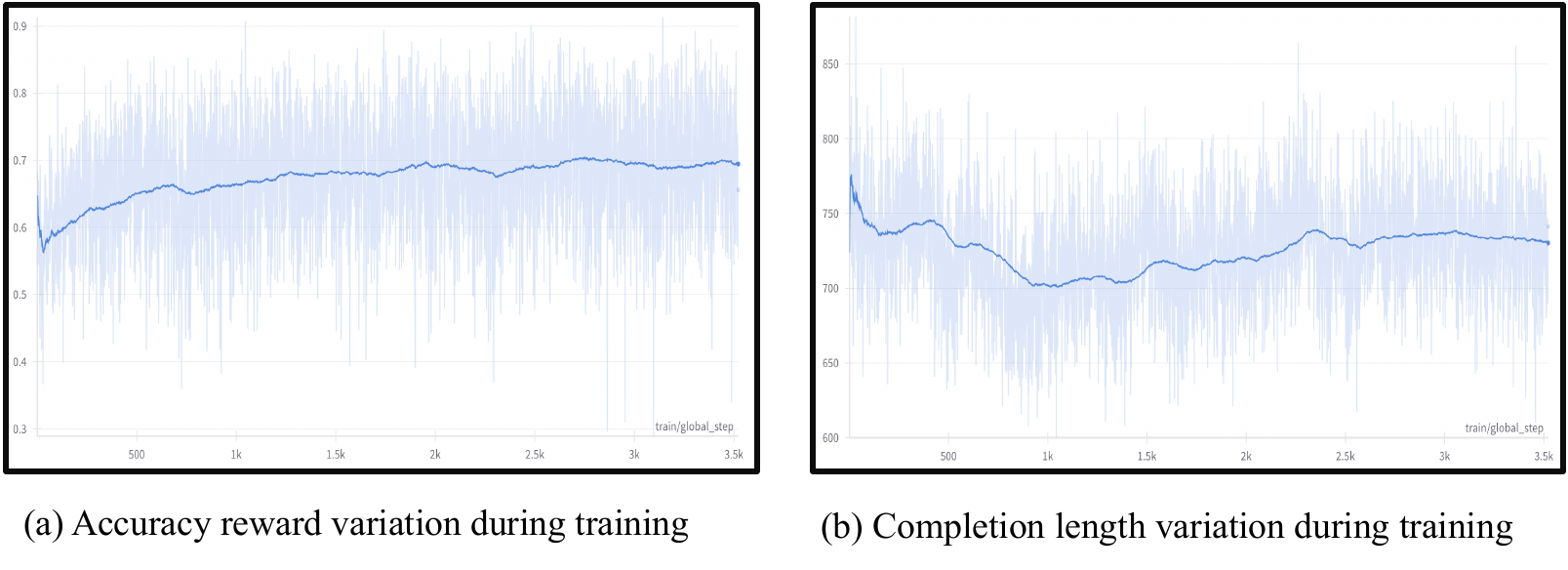}
    \caption{Training dynamics of Q-Ponder during GRPO fine-tuning. (a) Accuracy reward variation indicates a steady improvement trend in early stages, followed by stabilization. (b) Completion length initially maintains the full reasoning length distilled during the cold-start phase, then slightly decreases before gradually increasing again as part of a refinement process.}
    \label{fig5}
    \vspace{-5mm}
\end{figure}

\paragraph{Evidence of fast stabilization through cold-start initialization.}
In Fig. \ref{fig5}(a), the accuracy reward curve exhibits a rapid upward trend during the early phase of training, followed by a clear plateau. This pattern reflects an efficient transition from unstable, noisy outputs to reward-consistent reasoning. The absence of volatile oscillations after step ~800 suggests that the cold-start initialization (Q-Ponder-CI), distilled with expert reasoning chains, provides a strong starting policy that effectively narrows the exploration space for reinforcement learning. As a result, the GRPO policy optimization benefits from more stable trajectories and converges faster. This validates the hypothesis that cold-start supervision significantly accelerates convergence and improves training stability by offering semantically meaningful priors.

\paragraph{Improvement in reasoning compactness through reward-guided optimization.}
Figure \ref{fig5}(b) shows the evolution of the generated reasoning length during training. Initially, the model reproduces long reasoning chains inherited from the distillation stage, which reflects the verbose expert-generated format. However, during reinforcement fine-tuning, the completion length gradually declines—indicating that the model learns to avoid redundant reasoning segments. This suggests that the reward model not only guides score accuracy but also implicitly incentivizes more concise and goal-directed reasoning, likely because excessive length increases the risk of hallucinations and format violations (as penalized in $R_{\text{fmt}}$). Toward the final phase of training, we observe a slight increase in output length, which we interpret as the model refining its balance between informativeness and brevity. This aligns with the intended goal of optimizing both score alignment and reasoning quality in a structured manner.

\paragraph{Conclusion.}
Together, these two trends illustrate that our unified training pipeline enables Q-Ponder to transition from verbose, noisy reasoning toward compact, accurate, and consistent quality assessment. The cold-start stage provides a robust initialization, while the GRPO fine-tuning further optimizes both reasoning precision and structural quality in an efficient and interpretable manner.

\section{Qualitative Case Studies}
\label{D}
\paragraph{Overview.}
To further demonstrate the reasoning capabilities of different models, we conduct qualitative comparisons across \textbf{six models} on \textbf{three types of image sources}. Each case showcases the model's ability to interpret image quality through textual reasoning, highlighting both strengths and common pitfalls.

\vspace{-2mm}
\paragraph{Image Sources.}
We select representative images from the following domains:
\begin{itemize}
    \item \textbf{Image Restoration Datasets:} Contain synthetic degradations such as noise, blur, or compression artifacts.
    \item \textbf{Web Images:} Real-world images collected online, with diverse content and mixed quality.
    \item \textbf{Natural Photography:} High-quality landscape or portrait photographs, often with subtle degradations.
\end{itemize}
The ground truth score of the image is derived from human annotations, original distortion labels, or OpenAI O3 annotations.

\vspace{-2mm}
\paragraph{Compared Models.}
We include six representative models:
\begin{itemize}
    \item \textbf{Q-Ponder} (ours, after GRPO fine-tuning)
    \item \textbf{Q-Ponder-CI} (ours, cold-start only)
    \item \textbf{Qwen-2.5VL-7B} (base model)
    \item \textbf{DepictQA}
    \item \textbf{Co-Instruct}
    \item \textbf{Q-Instruct}
\end{itemize}

\vspace{-2mm}
\paragraph{Annotation Strategy.}
To better highlight model performance in reasoning:
\begin{itemize}
    \item \textcolor{blue}{\textbf{Blue}} is used to mark \textbf{correct judgments}, such as accurate identification of distortions or appropriate severity assessment.
    \item \textcolor{red}{\textbf{Red}} is used to mark \textbf{hallucinations or errors}, including incorrect distortion types, wrong scene interpretation, or overclaims.
\end{itemize}

\begin{figure}[htbp]
    \centering
    \includegraphics[width=0.83\textwidth]{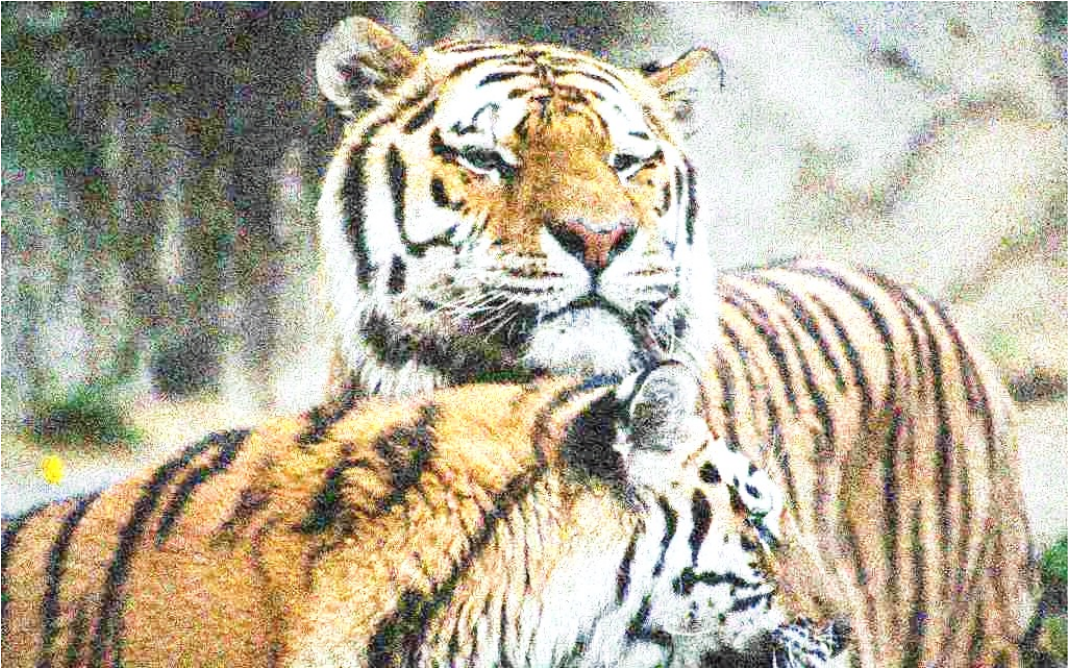} 
    \vspace{2pt}
    
    {\textbf{Case.1} Ground Truth Score: 26.35} 
    \vspace{-9mm}
\end{figure}

\renewcommand{\arraystretch}{1.4}
\setlength{\tabcolsep}{8pt}

\begin{small} 
\centering
\setlength\LTleft{-1in}
\setlength\LTright{-1in plus 1 fill}
\setlength{\tabcolsep}{2pt}
\begin{longtable}{|p{1.35\linewidth}|}
\caption*{} \\[-1.5ex] 

\hline
\multicolumn{1}{|c|}{\textbf{Q-Ponder}} \\
\hline

(1) In the low-level attribute analysis, the image exhibits significant issues in several key areas. Firstly, \textcolor{blue}{the brightness and exposure are problematic due to severe overexposure}, \textcolor{blue}{which causes a loss of detail in the highlights}, particularly \textcolor{blue}{around the tiger's fur and surrounding areas}. This results in an \textcolor{blue}{overall appearance where the image appears overly bright and lacks balance}. Secondly, \textcolor{blue}{contrast is severely compromised}; the differentiation between \textcolor{blue}{light and dark areas is minimal}, leading to a flat image where details are hard to distinguish. The \textcolor{blue}{global sharpness and detail preservation are gravely affected by the presence of extreme noise}, which obscures fine details and creates a grainy texture throughout the image. This \textcolor{blue}{noise appears to be a combination of digital noise and possibly compression artifacts}, further degrading the clarity. Additionally, \textcolor{blue}{color performance is distorted}, with \textcolor{blue}{unnatural color saturation and shifts, likely due to the noise and exposure issues}. The \textcolor{blue}{dynamic range is limited}, failing to preserve details in both dark and bright areas. There are no apparent lens distortions or chromatic aberrations, but the visual interference caused by the noise and overexposure is substantial, disrupting the depiction of the main subject and overall visual perception.
\newline
\\
(2) In the high-level attribute analysis, subject clarity is affected by the \textcolor{blue}{aforementioned noise and overexposure}, making the tiger less defined and its features harder to distinguish. \textcolor{blue}{The composition is straightforward, placing the tiger centrally}, but the lack of focus on the subject's details detracts from its effectiveness. There is little use of leading lines or balance, and the \textcolor{blue}{image lacks emotional storytelling due to the technical flaws}. The artistic style cannot be confidently identified because the \textcolor{blue}{image quality obscures any intentional creative choices}. The environment and background do not effectively complement the subject due to the \textcolor{blue}{lack of detail and context provided by the poor image quality}.
\newline
\\
(3) Considering the analysis of both low-level and high-level attributes, the image quality is \textcolor{blue}{significantly degraded}. To improve it, \textcolor{blue}{reducing exposure and adjusting contrast} would help recover lost details and create more natural-looking tones. Noise reduction techniques could enhance sharpness and clarity, bringing out finer textures in the tiger's fur. \textcolor{blue}{Adjusting color saturation and ensuring smoother transitions would improve realism}. Cropping or reframing might help focus attention on the subject, enhancing the composition. Addressing the technical issues would allow the inherent artistic and emotional qualities of the image to be better appreciated.
\newline
\\
<answer>\textcolor{blue}{24.8974}</answer> \\
\hline

\multicolumn{1}{|c|}{\textbf{Qwen2.5VL-7B}} \\
\hline

To evaluate the quality of the image, let's consider the following factors:

1. Clarity: The image appears to be \textcolor{red}{somewhat blurry}, which affects the overall clarity. The details of the tiger's fur and facial features are not very sharp.

2. Focus: \textcolor{red}{The focus seems to be off}, as the image \textcolor{red}{lacks sharpness in both the foreground and background}. This makes it difficult to discern fine details.

3. Lighting: The \textcolor{red}{lighting is adequate}, but there \textcolor{blue}{might be some overexposure} in certain areas, particularly around the tiger's face, which could be due to the bright background.

4. Color: \textcolor{red}{The colors appear natural}, but the image has a \textcolor{blue}{high level of noise or graininess}, which can be distracting and reduce the overall quality.

5. Composition: The \textcolor{blue}{composition is centered, which is acceptable}, but the lack of sharpness detracts from the visual impact.

6. Noise/Gain: There is \textcolor{blue}{noticeable noise or gain in the image}, which is evident in the grainy texture, especially in the darker areas.

\vspace{1em}
Considering these factors, the image quality is not high. It suffers from \textcolor{red}{blurriness}, \textcolor{red}{poor focus}, and \textcolor{blue}{excessive noise}, which significantly degrade the visual experience.

\vspace{1em}
Image Quality Score: \textcolor{red}{45.00}
\\
\hline

\multicolumn{1}{|c|}{\textbf{Q-Ponder-CI}} \\
\hline
1. In the analysis of low-level attributes, The image's \textcolor{blue}{brightness and exposure appear uneven}. There is a significant portion of the image that seems \textcolor{blue}{overexposed}, particularly in the lighter areas like the tiger's fur and background, resulting in a loss of detail. Conversely, some darker regions \textcolor{blue}{retain some texture but may lack depth due to the high brightness}. \textcolor{blue}{Contrast is excessively high}, creating a very sharp distinction between light and dark areas, which leads to a loss of subtlety in mid-tones. \textcolor{blue}{The global sharpness and detail preservation are compromised}, likely due to \textcolor{blue}{heavy compression or inherent image noise}, as \textcolor{blue}{evidenced by the grainy texture} throughout the image. The \textcolor{blue}{noise level is quite severe}, consisting of both digital noise and possibly compression artifacts, \textcolor{blue}{which obscure finer details} and make the image look less clear. \textcolor{blue}{Color performance is inconsistent}; while the \textcolor{red}{white balance appears neutral, the saturation is unnaturally high in certain areas}, especially around the tiger, leading to a lack of realistic color transitions. \textcolor{blue}{The dynamic range is limited}, with \textcolor{blue}{a significant loss of detail in both shadows and highlights}. No apparent lens distortion, chromatic aberration, or vignetting is present, but the image \textcolor{red}{suffers from glare or possible lens flare}, adding to the visual interference.
\newline
\\
2. In the analysis of high-level attributes, the subject clarity is compromised as the tiger, although central, lacks sharpness and detail due to the noise and overexposure. The composition and layout \textcolor{blue}{do not effectively utilize established principles like the rule of thirds or leading lines}; instead, the \textcolor{red}{chaos introduced by the noise disrupts any potential harmony}. \textcolor{blue}{Emotional expression and storytelling are hindered by the technical flaws}, making it difficult to convey any intended mood or narrative. There is little evidence of a deliberate artistic style beyond the technical issues present, though the \textcolor{red}{noise could unintentionally suggest a abstract or impressionist effect}. The integration of the environment and background is poor, as the \textcolor{blue}{background details are lost in the overexposure and noise}, failing to complement the subject. Lastly, context adaptability is lacking since the image does not meet the basic quality standards expected for wildlife photography, where detail and natural color reproduction are paramount.
\newline
\\
3. Comprehensively evaluating the image, it suffers heavily from technical faults that overshadow any potential aesthetic qualities. The \textcolor{blue}{severe denoising distortion, overexposure, and loss of detail} significantly degrade the image quality. Improvements should focus on adjusting exposure settings to avoid overexposure, reducing noise through \textcolor{blue}{better lighting conditions or post-processing techniques}, and ensuring proper contrast for a more balanced and detailed image. Attention to these low-level attributes will enhance high-level aspects such as subject clarity, composition, and emotional impact. Given the extent of the issues, a re-shoot under controlled conditions might be necessary to achieve a higher quality image.
\newline
\\
<answer>\textcolor{blue}{18.5478}</answer>
\\
\hline

\multicolumn{1}{|c|}{\textbf{DepictQA}} \\
\hline
The image depicts a \textcolor{blue}{lion} with a distinctive mane. The evaluated image \textcolor{red}{suffers from serious noise}, evident in the \textcolor{blue}{grainy texture} that obscures fine details, and \textcolor{red}{serious compression artifacts}, which manifest as \textcolor{red}{blocky regions and loss of detail, particularly noticeable in the lion's mane and face}. These distortions significantly degrade the visual clarity, making the image appear less sharp and the subject less recognizable. Overall, the quality of the evaluated image is \textcolor{blue}{poor}, with the distortions heavily impacting the viewer's ability to appreciate the content and details of the original image.\\
\hline

\multicolumn{1}{|c|}{\textbf{Co-Instruct}} \\
\hline
The overall clarity of this image is low, with \textcolor{red}{weak lighting} and \textcolor{blue}{some noise present}. The texture details of the \textcolor{blue}{tiger}, the main subject, are \textcolor{blue}{partially lost and slightly blurred}. The composition is not good, with a \textcolor{red}{slight tilt}. The background is \textcolor{red}{somewhat blurry and has lost most of its texture details}. Therefore, the quality of this image is poor. \\
\hline

\multicolumn{1}{|c|}{\textbf{Q-Instruct}} \\
\hline
The overall clarity of this image is very low, with \textcolor{red}{weak lighting}. The main subject, a \textcolor{blue}{tiger}, has lost most of its \textcolor{blue}{texture details and appears blurry}. The composition is poor, with a \textcolor{red}{slight tilt}, and the \textcolor{red}{background is also blurry}. Therefore, the image quality of this photo is very poor.
 \\
\hline

\end{longtable}
\end{small}

\newpage

\begin{figure}[htbp]
    \centering
    \includegraphics[width=0.54\textwidth]{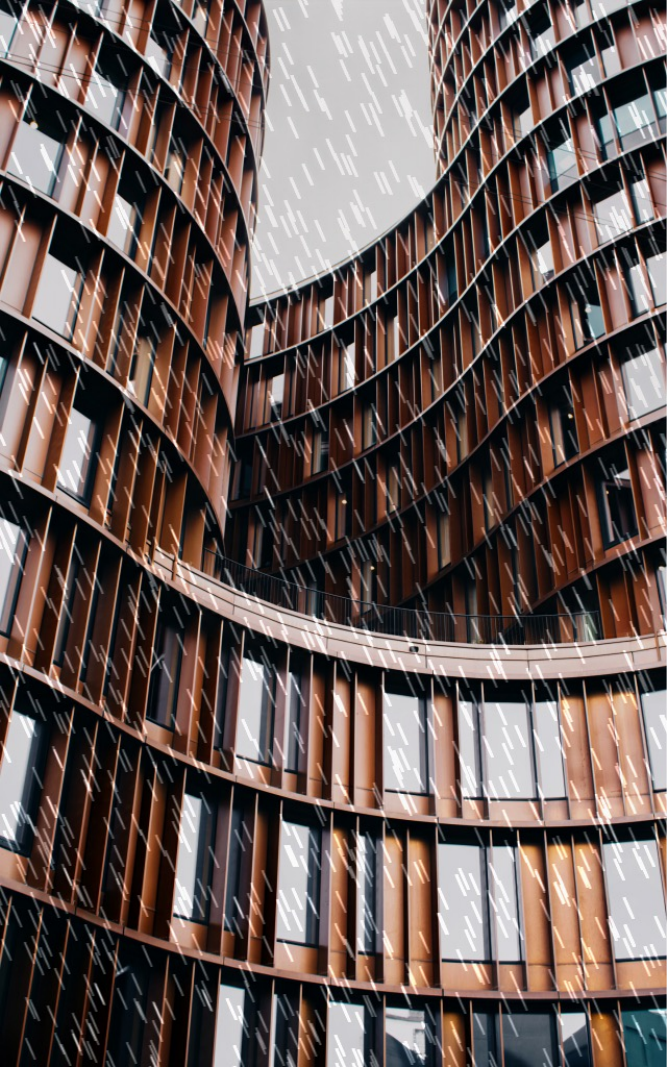} 
    \vspace{2pt}
    
    {\textbf{Case2} Ground Truth Score: 45.00} 
    \vspace{-9mm}
\end{figure}

\renewcommand{\arraystretch}{1.4}
\setlength{\tabcolsep}{8pt}

\begin{small} 
\centering
\setlength\LTleft{-1in}
\setlength\LTright{-1in plus 1 fill}
\setlength{\tabcolsep}{2pt}
\begin{longtable}{|p{1.35\linewidth}|}
\caption*{} \\[-1.5ex] 

\hline
\multicolumn{1}{|c|}{\textbf{Q-Ponder}} \\
\hline

(1) In the low-level attribute analysis, the \textcolor{blue}{brightness and exposure of the image appear moderately balanced}, with \textcolor{red}{neither underexposure nor overexposure} dominating the scene. The \textcolor{blue}{lighting is natural}, with sunlight filtering through the windows, creating a play of light and shadow that enriches the image's depth. The \textcolor{blue}{contrast is well-managed}, with distinct differences between illuminated and shaded areas enhancing the architectural details \textcolor{blue}{without causing any significant loss of detail in either highlights or shadows}. The \textcolor{blue}{global sharpness and detail preservation are commendable}, as the \textcolor{blue}{intricate patterns of the building's facade and the texture of the materials are clearly visible}. However, there is a \textcolor{blue}{noticeable level of noise present}, likely due to the \textcolor{blue}{processing limitations}, which \textcolor{blue}{affects the clarity}. The \textcolor{blue}{color performance is consistent}, with \textcolor{blue}{accurate white balance and natural color saturation that reflect the true tones of the structure}. The \textcolor{red}{dynamic range is adequate, preserving details in both bright and dark regions}. Nevertheless, \textcolor{blue}{there is a form of distortion—slant lines distortion—visible in the image, likely introduced during digital processing, which alters the architectural lines slightly}. There are no apparent lens distortions, chromatic aberrations, or vignetting. \textcolor{blue}{Special visual interference factors include slant lines, which may be artificially added rain-like noise. These lines can distract from the overall architectural aesthetics and degrade the image quality to some extent}.

\\
(2) In the high-level attribute analysis, \textcolor{blue}{the subject clarity is high, with the building's architectural features being the focal point and standing out distinctly against the background}. The \textcolor{blue}{composition and layout utilize leading lines effectively}, drawing the viewer's eye upwards along the curved architecture, \textcolor{blue}{though the slant lines introduce a disruption in this flow}. The image \textcolor{blue}{does not strongly convey emotional expression or storytelling}, as it focuses more on \textcolor{blue}{architectural documentation rather than evoking deep emotions}. The artistic style leans towards \textcolor{blue}{modern architecture}, with \textcolor{blue}{clean lines and a structured approach}, yet the distortion slightly detracts from this aesthetic. The integration of environment and background is harmonious, with no distracting elements, \textcolor{blue}{except for the slant lines}. The \textcolor{blue}{context adaptability is appropriate for architectural photography}, emphasizing \textcolor{blue}{structural details and design}.
\newline
\\
(3) Comprehensively evaluating the image, its \textcolor{blue}{quality is relatively good}, with \textcolor{blue}{strengths in exposure, contrast, sharpness, and color accuracy}. However, the \textcolor{blue}{presence of slant line distortion diminishes the overall quality to some extent}. To improve, \textcolor{blue}{reducing the noise level and correcting the distortion could enhance clarity and architectural authenticity}. Additionally, while the image excels in structural details, incorporating more elements to evoke an emotional response could elevate its artistic value. Considering all factors, the image demonstrates a solid foundation in technical attributes but \textcolor{blue}{has room for enhancement in noise reduction and distortion correction}.
\\
<answer>\textcolor{blue}{52.4316}</answer> \\
\hline

\multicolumn{1}{|c|}{\textbf{Qwen2.5VL-7B}} \\
\hline
To evaluate the quality of the image, I will consider several factors such as clarity, composition, lighting, color, and detail.

\vspace{1em}
\noindent
1. Clarity: \textcolor{red}{The image appears to be clear}, with \textcolor{red}{distinct lines and shapes} visible in the architectural structure. There is \textcolor{red}{no noticeable blurriness or distortion that would affect the overall clarity}.

\vspace{0.5em}
\noindent
2. Composition: The \textcolor{blue}{composition is well-balanced}, with the \textcolor{blue}{circular structure creating a strong visual interest}. The \textcolor{blue}{perspective used emphasizes the depth and curvature} of the building, making it engaging.

\vspace{0.5em}
\noindent
3. Lighting: The lighting is even, with natural light illuminating the facade of the building. There are \textcolor{red}{no harsh shadows or overexposed areas}, which suggests good exposure management.

\vspace{0.5em}
\noindent
4. Color: The \textcolor{blue}{colors are natural and well-saturated}, with a \textcolor{blue}{warm tone dominating due to the material of the building}. This adds to the aesthetic appeal of the image.

\vspace{0.5em}
\noindent
5. Detail: The details of the building's facade are sharp and well-defined, \textcolor{red}{allowing for a clear view of the architectural features}.

\vspace{1em}
\noindent
Based on these observations, the image quality is \textcolor{red}{high}. Here is the assigned score:\\
Score: \textcolor{red}{95.00}
\\
\hline

\multicolumn{1}{|c|}{\textbf{Q-Ponder-CI}} \\
\hline
1. The image's \textcolor{blue}{brightness and exposure seem appropriate for the scene}, capturing the interplay of light and shadow across the building's facade. However, some areas near the light source are \textcolor{blue}{slightly overexposed}, leading to minor detail loss. \textcolor{blue}{Contrast is well-managed}, providing a \textcolor{blue}{sense of depth and dimensionality to the structure}. \textcolor{blue}{Global sharpness and detail preservation are fairly good}, with clear architectural lines and textures visible. There is \textcolor{blue}{minimal noise}, suggesting either a low ISO setting was used or effective noise reduction during processing. \textcolor{blue}{Colors appear natural}, with \textcolor{blue}{accurate white balance and reasonable saturation levels}, though the \textcolor{red}{warm tones of the building's exterior give a consistent hue} throughout the image. \textcolor{blue}{The dynamic range is adequate, preserving details in both bright and shaded areas}. No significant lens distortion, chromatic aberration, or vignetting is evident, indicating a well-calibrated camera or lens setup. \textcolor{red}{Some visual interference may arise from the lighting conditions, casting shadows that interact with the building's geometry}
\\
2. The subject, which is the architectural structure, is \textcolor{blue}{clearly defined against the sky}. The composition employs \textcolor{blue}{leading lines created by the building's curved facade}, guiding the viewer's eye towards the center where the light is concentrated. This contributes to a \textcolor{blue}{balanced and engaging layout}. The image conveys a sense of modernity and architectural beauty, \textcolor{blue}{evoking an emotional response tied to urban design and engineering}. The artistic style leans toward \textcolor{blue}{contemporary architecture photography}, emphasizing form and structure. \textcolor{red}{The background integrates seamlessly with the subject, enhancing the overall visual experience without drawing attention away from the main focus}.
\\
3. Overall, \textcolor{red}{the image displays good quality with only minor issues in overexposure and shadow detail loss}. Improvements could involve fine-tuning exposure settings to prevent highlight clipping and \textcolor{blue}{using HDR techniques to recover shadow details}. \textcolor{blue}{Enhancing the contrast subtly can further define the architectural elements}. For the high-level attributes, \textcolor{blue}{maintaining the current composition and emotional storytelling is advisable}, as they contribute positively to the image's appeal. \textcolor{blue}{Minor adjustments in post-processing could optimize color balance and saturation for a more vibrant outcome}. With these refinements, the image would reach a higher standard of technical and aesthetic excellence.
\\
<answer>\textcolor{red}{75.6938}</answer>
\\
\hline

\multicolumn{1}{|c|}{\textbf{DepictQA}} \\
\hline
The images depict a \textcolor{blue}{modern building with a geometric facade}. In the evaluated image, \textcolor{red}{slight pixelation is observed, reducing the sharpness and detail of the architectural features}. Despite this, the overall structure and color remain recognizable. The image maintains a \textcolor{blue}{fair level of quality}, with the main subject still discernible, but \textcolor{red}{with a noticeable loss in fine details}.\\
\hline

\multicolumn{1}{|c|}{\textbf{Co-Instruct}} \\
\hline
The overall clarity of this image is acceptable, with \textcolor{blue}{sufficient lighting}. The texture details of the main subject, the building, \textcolor{red}{are rich and clear}. The colors are also rich. However, \textcolor{red}{the composition is not good, as it is slightly tilted}. The background is \textcolor{red}{somewhat blurry}, and \textcolor{red}{most of the texture details are almost lost}. Therefore, the image quality is good. \\
\hline

\multicolumn{1}{|c|}{\textbf{Q-Instruct}} \\
\hline
The overall clarity of this image is \textcolor{blue}{acceptable}, with \textcolor{blue}{sufficient lighting}. The \textcolor{red}{texture details of the main subject}, the building, \textcolor{red}{are rich and clear}. The colors are also rich. However, \textcolor{red}{the composition is not good, as it is slightly tilted}. The background is somewhat blurry. Therefore, the image quality is good.
 \\
\hline

\end{longtable}
\end{small}

\newpage

\begin{figure}[htbp]
    \centering
    \includegraphics[width=0.50\textwidth]{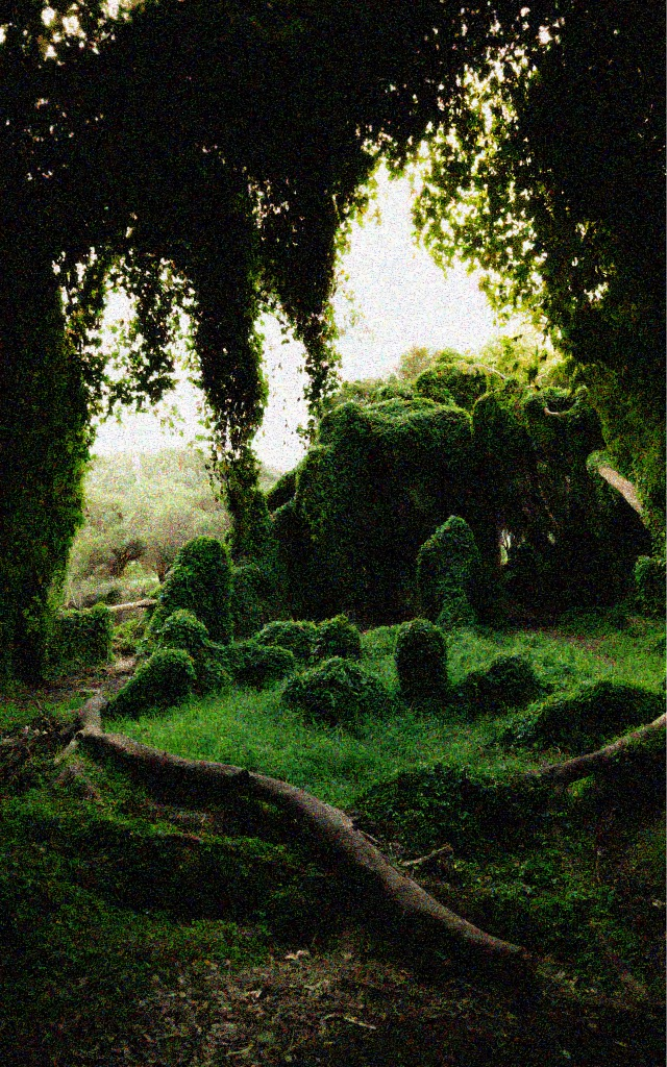} 
    \vspace{2pt}
    
    {\textbf{Case.3} Ground Truth Score: 30.14} 
    \vspace{-9mm}
\end{figure}

\renewcommand{\arraystretch}{1.4}
\setlength{\tabcolsep}{8pt}

\begin{small} 
\centering
\setlength\LTleft{-1in}
\setlength\LTright{-1in plus 1 fill}
\setlength{\tabcolsep}{2pt}
\begin{longtable}{|p{1.35\linewidth}|}
\caption*{} \\[-1.5ex] 

\hline
\multicolumn{1}{|c|}{\textbf{Q-Ponder}} \\
\hline

(1) In the low-level attribute analysis, the image \textcolor{blue}{exhibits significant issues in several key areas}. Firstly, the \textcolor{blue}{brightness and exposure are problematic}, with \textcolor{blue}{the upper portion of the image being overexposed}, leading to a \textcolor{blue}{loss of detail in the bright areas}. Conversely, \textcolor{blue}{the lower part appears underexposed}, resulting in a \textcolor{blue}{lack of visible details in darker regions}. This imbalance affects the overall exposure balance. The \textcolor{blue}{contrast is excessive}, causing a \textcolor{blue}{harsh separation between light and dark areas}, which further diminishes the detail preservation. The global sharpness and detail preservation are compromised due to the \textcolor{blue}{presence of severe noise throughout the image}, likely originating from the sensor due to high ISO settings or poor lighting conditions. \textcolor{blue}{This noise significantly degrades the clarity and sharpness of the image}. The \textcolor{blue}{color performance is inconsistent}, with an \textcolor{blue}{unnatural white balance} that skews the colors towards \textcolor{red}{a warm tone}. The \textcolor{blue}{color saturation appears artificial}, and there are \textcolor{red}{edge/transition color shifts} that disrupt the natural appearance. \textcolor{blue}{The dynamic range is limited, as both the dark and bright areas lack detail}, suggesting the sensor struggles with capturing a full spectrum of light. Distortion artifact detection reveal \textcolor{red}{the presence of chromatic aberration, particularly noticeable along the edges of the tree branches}. There are \textcolor{blue}{no evident lens flare or other visual interferences}, but the overall distortion affects the image's clarity and visual perception.
\newline
\\
(2) In the high-level attribute analysis, the \textcolor{blue}{subject clarity is affected by the noise and exposure issues}, \textcolor{blue}{making it difficult to distinguish the subject from the background}. The composition and layout \textcolor{blue}{attempt to use leading lines with the tree branches and the path}, but the \textcolor{blue}{excessive noise and distortion detract from this effect}. The emotional expression and storytelling potential are \textcolor{blue}{hindered by the technical flaws}, as \textcolor{blue}{the intended mood or narrative is obscured by the visual distortions}. There is \textcolor{blue}{no clear artistic style or creativity evident due to the overwhelming presence of noise and distortion}. The integration of the environment and background is poor, with the background elements not complementing the subject effectively. The \textcolor{blue}{context adaptability is also compromised}, as \textcolor{blue}{the image fails to meet the standards expected for landscape photography}, where detail preservation and dynamic range are crucial. The distortion and noise levels distract from the natural beauty that landscape photography aims to capture.
\newline
\\
(3) Comprehensively evaluating the image, the \textcolor{blue}{low-level attributes dominate the assessment, significantly impacting the overall quality}. The \textcolor{blue}{severe noise, distortion, exposure imbalance, and color inconsistencies severely degrade the image's technical quality}. The high-level attributes suffer as a consequence of these technical issues, failing to achieve the compositional and emotional impact that would enhance the image. Improvement suggestions include \textcolor{blue}{reducing noise through better lighting or lower ISO settings}, \textcolor{blue}{adjusting exposure to balance the highlights and shadows}, and \textcolor{blue}{refining color balance for a more natural appearance}. Post-processing techniques could help enhance detail and reduce noise, while ensuring the composition is clearer and more engaging. Given the extent of the issues, the image requires substantial improvement in both technical and aesthetic aspects.
\newline
\\
<answer>\textcolor{blue}{23.6641}</answer>  \\
\hline

\multicolumn{1}{|c|}{\textbf{Qwen2.5VL-7B}} \\
\hline
To evaluate the quality of the image, let's consider several factors:

\noindent
1. Clarity and Focus: The image \textcolor{red}{appears somewhat blurry, which suggests that it may not be in sharp focus}. This can make it difficult to discern fine details.

\noindent
2. Lighting: \textcolor{red}{The lighting seems natural} but is \textcolor{blue}{somewhat overexposed in certain areas, particularly around the center where the light is brightest}. This can \textcolor{red}{wash out details and create glare}.

\noindent
3. Color Saturation: \textcolor{red}{The colors appear quite vivid, especially the greens}, which might be due to the natural environment or post-processing effects. However, \textcolor{blue}{the saturation could be adjusted for better balance}.

\noindent
4. Composition: The \textcolor{blue}{composition is interesting, with a natural frame created by the trees and a path leading into the scene}. However, the overexposure in the center might distract from the overall composition.

\noindent
5. Noise and Grain: There \textcolor{blue}{appears to be some noise or graininess, particularly noticeable in the darker areas, which can reduce the perceived quality of the image}.

\noindent
6. Detail Retention: Due to the \textcolor{red}{blurriness} and \textcolor{blue}{overexposure}, there is a loss of detail in the image, making it harder to appreciate finer elements of the scene.

Considering these points, here is a step-by-step evaluation:

  Clarity and Focus: 30/100 \textcolor{blue}{(low clarity)}
  Lighting: 40/100 \textcolor{blue}{(overexposed in parts)}
  Color Saturation: 70/100 \textcolor{red}{(vivid but potentially too high)}
  Composition: 60/100 \textcolor{blue}{(interesting but compromised by overexposure)}
  Noise and Grain: 40/100 \textcolor{red}{(moderate noise)}

\noindent
Overall Quality Score: \textcolor{red}{48.00}
\\
\hline
\multicolumn{1}{|c|}{\textbf{Q-Ponder-CI}} \\
\hline
1. The image's \textcolor{blue}{brightness and exposure show a mix of challenges}. \textcolor{blue}{The brighter areas near the top where sunlight filters through the trees seem slightly overexposed}, causing some loss of detail. Conversely, \textcolor{blue}{the lower parts of the image are underexposed}, resulting in a \textcolor{blue}{lack of visibility in the darker regions}. \textcolor{blue}{The contrast is quite pronounced}, with a \textcolor{blue}{stark difference between the illuminated paths and the shaded areas beneath the trees}. This high contrast creates a dramatic effect but also introduces a risk of losing fine details in both highlights and shadows. \textcolor{red}{Global sharpness is compromised, likely due to a combination of camera shake, focus issues, or post-processing effects}, as indicated by the \textcolor{red}{noticeable graininess and lack of crisp edges on objects like branches and foliage}. \textcolor{blue}{The noise level is elevated, manifesting as digital noise, possibly from a high ISO setting or compression artifacts, which degrade the overall clarity}. \textcolor{blue}{Color performance is somewhat inconsistent}, with an imbalance in white balance leaning towards \textcolor{red}{cooler tones in the shaded areas and warmer tones in the sunlit regions}. This inconsistency affects color fidelity and smoothness. \textcolor{blue}{The dynamic range is limited}, struggling to capture details across the spectrum of light and dark areas. There is no apparent lens distortion, chromatic aberration, or vignetting, but the image \textcolor{red}{contains visual interference such as glare or lens flare, likely caused by direct sunlight hitting the camera sensor}.\newline
\\
2. \textcolor{blue}{Subject clarity is impacted by the lack of sharpness and detail, making it difficult to distinguish specific elements within the scene}. The composition uses \textcolor{blue}{leading lines formed by the tree trunks and branches}, subtly guiding the viewer's gaze through the image. However, \textcolor{blue}{the lack of focus and detail reduces the effectiveness of this technique}. The image \textcolor{blue}{struggles to convey a strong emotional narrative}, partly because of the technical issues that overshadow the potential mood set by the dappled light and shadow. \textcolor{blue}{The artistic style is hard to pinpoint}, but the image has a \textcolor{blue}{raw, almost abstract quality} due to the interplay of light, shadow, and color. \textcolor{blue}{The integration of the environment and background is decent}, with the forest floor and vegetation providing context, but \textcolor{blue}{the lack of clarity hinders a harmonious blend}
\newline
\\
3. Considering all aspects, the image has \textcolor{red}{moderate quality} with several areas needing improvement. \textcolor{blue}{Reducing exposure in the brighter spots and increasing fill light in the darker areas} would help achieve a more balanced exposure. \textcolor{blue}{Adjusting contrast and sharpening the image could enhance detail and clarity}. Noise reduction techniques would \textcolor{blue}{clean up the grainy texture}, improving overall image quality. \textcolor{blue}{Fine-tuning the white balance could produce more natural and consistent colors}. Cropping or reframing might help emphasize key elements, strengthening the composition. Addressing the technical issues would allow the image's artistic and emotional qualities to shine through more effectively.
\newline
\\
<answer>\textcolor{red}{45.5473}</answer>
\\
\hline

\multicolumn{1}{|c|}{\textbf{DepictQA}} \\
\hline
The image depicts a \textcolor{blue}{natural scene with vegetation and a path}. The evaluated image \textcolor{blue}{exhibits significant noise}, which \textcolor{blue}{obscures fine details and creates a grainy appearance}. Additionally, the colors appear washed out, \textcolor{red}{indicating a reduction in saturation}, which \textcolor{red}{diminishes the vibrancy and contrast of the image}. Overall, the image quality is \textcolor{blue}{compromised}, with the \textcolor{blue}{noise severely affecting clarity} and the \textcolor{blue}{desaturation} impacting the visual appeal, resulting in a less engaging and detailed representation of the scene.\\
\hline

\multicolumn{1}{|c|}{\textbf{Co-Instruct}} \\
\hline
The \textcolor{blue}{overall clarity of this image is low}, with \textcolor{blue}{weak lighting and some noise present}. The green plants, which are the main subject, have \textcolor{blue}{lost some texture details} and appear \textcolor{red}{slightly blurry}. The \textcolor{red}{composition is not good, with a slight tilt}. The background is also \textcolor{red}{somewhat blurry}, with the loss of most texture details. Therefore, \textcolor{blue}{the quality of this image is poor}. \\
\hline

\multicolumn{1}{|c|}{\textbf{Q-Instruct}} \\
\hline
The overall clarity of this image is \textcolor{blue}{very low}, with \textcolor{blue}{weak lighting}. The main subject, the trees, have lost most of their texture details. There is \textcolor{blue}{some noise present}, and the \textcolor{red}{composition is poor, with a slight tilt}. The \textcolor{red}{background is also blurry}. Therefore, \textcolor{blue}{the quality of this image is very poor}.
 \\
\hline

\end{longtable}
\end{small}

\newpage

\begin{figure}[htbp]
    \centering
    \includegraphics[width=0.82\textwidth]{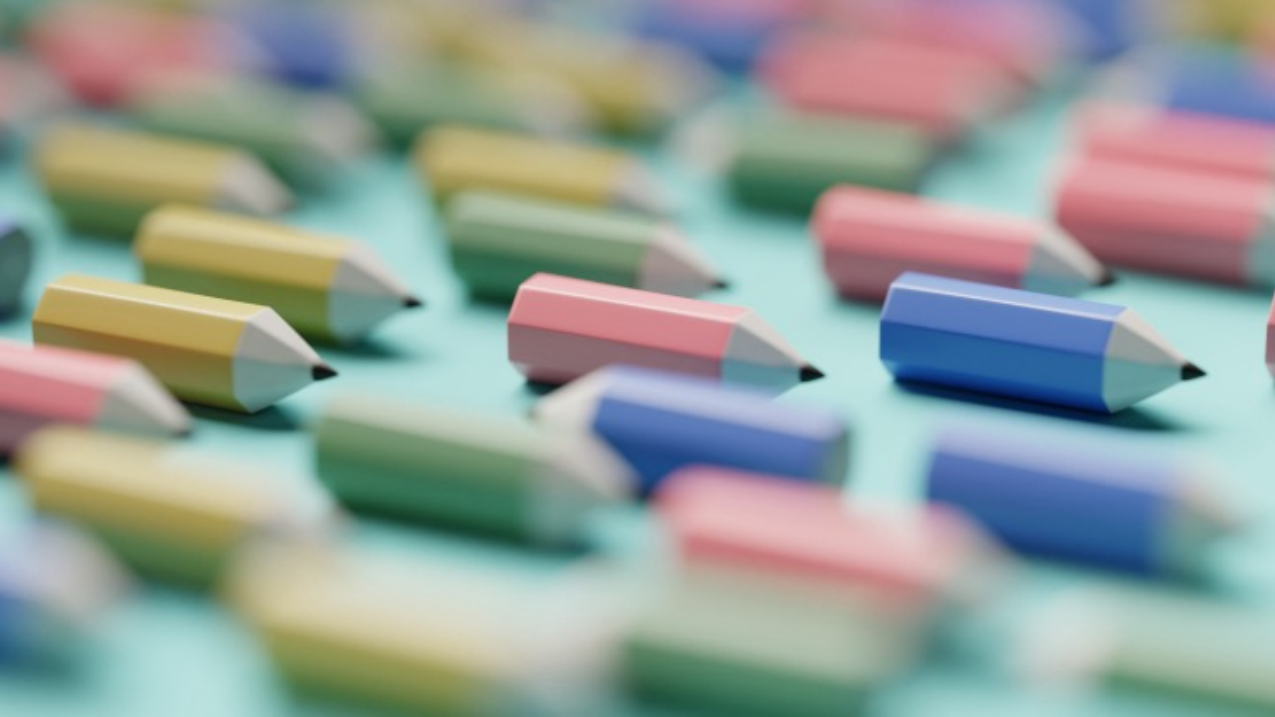} 
    \vspace{2pt}
    
    {\textbf{Case.4} Ground Truth Score: 92.00} 
    \vspace{-7mm}
\end{figure}

\renewcommand{\arraystretch}{1.4}
\setlength{\tabcolsep}{8pt}

\begin{small} 
\centering
\setlength\LTleft{-1in}
\setlength\LTright{-1in plus 1 fill}
\setlength{\tabcolsep}{2pt}
\begin{longtable}{|p{1.35\linewidth}|}
\caption*{} \\[-1.5ex] 

\hline
\multicolumn{1}{|c|}{\textbf{Q-Ponder}} \\
\hline

(1) In the low-level attribute analysis, the image \textcolor{blue}{demonstrates well-balanced brightness and exposure}, with \textcolor{blue}{neither underexposure nor overexposure} negatively affecting the visibility of details. The lighting is \textcolor{blue}{evenly distributed}, ensuring that \textcolor{blue}{all pencils are adequately illuminated}. The \textcolor{blue}{contrast is moderate}, providing \textcolor{blue}{sufficient differentiation between light and dark areas to maintain depth and dimensionality}. The image shows \textcolor{blue}{no significant noise} or \textcolor{red}{blurring}, preserving a \textcolor{red}{high level of clarity and sharpness}. \textcolor{blue}{Color performance is satisfactory}, with \textcolor{blue}{accurate white balance and moderate saturation}. \textcolor{blue}{The pastel tones of the pencils appear vibrant yet not overly saturated}, contributing to a \textcolor{blue}{pleasant visual effect}. \textcolor{blue}{The dynamic range is adequate}, capturing details in both highlights and shadows. No noticeable lens distortion, chromatic aberration, or vignetting is detected. There are no apparent visual distractions or occlusions, and overall, \textcolor{blue}{the image exhibits excellent visual quality}.
\newline
\\
(2) In the high-level attribute analysis, the \textcolor{blue}{subject clarity is acceptable}, with the \textcolor{blue}{focal point on the foreground pencils being relatively clear and distinct from the blurred background}. The \textcolor{blue}{shallow depth of field successfully isolates the main subjects, drawing attention to their shapes and colors}. The composition utilizes a \textcolor{blue}{shallow depth of field creatively}, employing \textcolor{blue}{bokeh to separate the subject from the background}. The arrangement of the pencils \textcolor{blue}{follows a loose pattern}, creating a \textcolor{blue}{sense of movement and dynamism across the frame}. While the image does not strictly adhere to traditional compositional rules like the golden ratio, the scattered placement of the pencils provides a \textcolor{blue}{casual and engaging visual flow}. The image \textcolor{blue}{conveys a playful and creative emotional expression} through its use of vibrant colors and the whimsical nature of the subject matter. The artistic style is \textcolor{blue}{minimalist and modern}, focusing on simplicity and color to create a visually appealing scene. The \textcolor{blue}{background complements the subject well}, with the uniform blue surface providing a clean and unobtrusive backdrop that \textcolor{blue}{enhances the colors of the pencils}.
\newline
\\
(3) In the overall assessment, the image stands out as a \textcolor{blue}{high-quality visual sample}, demonstrating \textcolor{blue}{strong performance in both technical and aesthetic dimensions}. Its balanced exposure, accurate colors, and clean composition contribute to an appealing visual experience. While the image already exhibits excellent clarity and minimal distractions, \textcolor{blue}{slight enhancements could further refine its presentation}. Applying targeted noise reduction in darker regions may help reveal more fine details, and a subtle increase in contrast could reinforce subject separation and depth. Improving the focus precision on the foreground pencils and minimizing the surface glare would polish the image even further. These adjustments are minor, \textcolor{blue}{as the image in its current form already reflects a high standard of quality}.
\newline
\\
<answer>\textcolor{blue}{89.0312}</answer>  \\
\hline

\multicolumn{1}{|c|}{\textbf{Qwen2.5VL-7B}} \\
\hline
To evaluate the quality of this image, I will consider several key factors: clarity, focus, color, composition, and noise.

1. Clarity: \textcolor{blue}{The image appears to be clear, with distinct edges and shapes visible in the pencils. The details are sharp, especially in the foreground where the pencils are in focus.}

2. Focus: \textcolor{blue}{The image is focused on the pencils in the foreground}, while \textcolor{blue}{the background is intentionally blurred (bokeh effect)}. This creates a pleasing depth of field that \textcolor{blue}{draws attention to the main subject}.

3. Color: \textcolor{blue}{The colors are vibrant and well-saturated}, contributing to the overall appeal of the image. \textcolor{blue}{The pastel shades of the pencils are visually appealing and add to the aesthetic quality}.

4. Composition: The composition is \textcolor{blue}{simple yet effective}. The arrangement of the pencils in the foreground and the blurred background \textcolor{blue}{creates a sense of depth and balance}. \textcolor{blue}{The angle at which the pencils} are placed adds visual interest.

5. Noise: There is \textcolor{blue}{no noticeable noise or graininess in the image}, indicating good image quality.

6. Overall Impression: The image is \textcolor{blue}{aesthetically pleasing and effectively conveys its subject matter}. The use of bokeh and the vibrant colors enhance the visual impact.

Based on these observations, I would assign the image a quality score of \textcolor{blue}{95.00}.
\\
\hline
\multicolumn{1}{|c|}{\textbf{Q-Ponder-CI}} \\
\hline
1.In the low-level attribute analysis, the image exhibits \textcolor{red}{significant issues primarily related to sharpness and detail preservation}. The majority of the image appears blurry, indicating \textcolor{red}{possible motion blur or focus issues}, which \textcolor{red}{directly impacts the clarity of the colored pencils}. This blurriness leads to a lack of fine detail presentation. Additionally, \textcolor{blue}{the contrast seems moderate}, but \textcolor{red}{due to the blur, the differentiation between light and dark areas is not pronounced}, resulting in a somewhat flat appearance. The global sharpness is compromised, and there is a \textcolor{red}{noticeable absence of crisp edges}. \textcolor{red}{The noise level is difficult to determine} precisely due to the blur, but any potential noise is likely overshadowed by the dominant blurriness. \textcolor{blue}{The color performance shows a wide range of pastel hues}, but \textcolor{red}{the color fidelity and saturation cannot be fully assessed because of the lack of sharpness}, making it hard to discern true colors versus what the blur introduces. \textcolor{blue}{There is no evident issue with dynamic range} since the scene is well-lit overall, but \textcolor{red}{the lack of sharpness prevents the full appreciation of details in both dark and bright areas}. No specific distortion like chromatic aberration or lens flare is apparent, but the blur itself acts as a form of distortion. There are no visible artifacts or moiré patterns. However, \textcolor{red}{the strong visual interference caused by the blur disrupts the overall visual perception and detracts from the image's quality}.
\newline
\\
2. Moving to the high-level attribute analysis, the subject clarity suffers significantly due to the blurriness, \textcolor{red}{making it challenging to distinguish the main subject from the background}. The composition and layout \textcolor{blue}{appear to aim for a scattered, casual arrangement of the pencils}, but the \textcolor{red}{lack of sharpness hinders the effectiveness of this intended randomness}. The image \textcolor{blue}{does not strongly convey emotion or storytelling} due to the primary focus being on the subject matter rather than an emotional narrative. There is \textcolor{red}{no distinct artistic style or creativity showcased beyond the choice of pastel colors}, which are somewhat obscured by the blur. \textcolor{red}{The integration of the environment and background is poor} because the background fails to complement the subject due to the extensive blur. The context adaptability is limited since \textcolor{red}{the image does not meet the basic requirements for a clear still life photograph}, where sharpness and detail are crucial.
\newline
\\
3. Comprehensively evaluating the image quality, the \textcolor{red}{severe global blur has a substantial negative impact across both low-level and high-level attributes}. While the color palette offers a pleasant pastel aesthetic, the lack of sharpness overshadows this positive aspect, leading to an overall diminished image quality. To improve the image, \textcolor{blue}{ensuring proper focus and minimizing motion during capture would enhance sharpness and detail}. Adjusting the lighting to create more contrast could also help in bringing out the features of the colored pencils. Revisiting the composition to emphasize specific pencils with better focus and lighting could elevate the artistic intent. Given the current state, \textcolor{red}{the image struggles to achieve its potential due to the blur, affecting its ability to effectively communicate visually}.
\newline
\\
<answer>\textcolor{red}{29.7710}</answer>
\\
\hline

\multicolumn{1}{|c|}{\textbf{DepictQA}} \\
\hline
The image depicts a \textcolor{red}{colorful pattern with a mix of blue and yellow hues}. The evaluated image suffers from \textcolor{red}{severe pixelation}, causing a \textcolor{red}{significant loss of detail and making the pattern appear blocky and less recognizable}. Additionally, the \textcolor{red}{contrast appears to be slightly reduced}, resulting in a flatter appearance with less visual depth. Overall, \textcolor{red}{the quality of the evaluated image is poor} due to the \textcolor{red}{heavy pixelation} which dominates the visual experience, while the slight contrast weakening further diminishes the image's clarity and vibrancy.\\
\hline

\multicolumn{1}{|c|}{\textbf{Co-Instruct}} \\
\hline
The \textcolor{red}{overall clarity of this image is acceptable}. \textcolor{blue}{The main subject, the pencil tip, retains most of its texture details}. The background, which is a table, \textcolor{blue}{is slightly blurry and exhibits some noise}. Therefore, the quality of this image \textcolor{red}{is considered average}. \\
\hline

\multicolumn{1}{|c|}{\textbf{Q-Instruct}} \\
\hline
The \textcolor{red}{overall clarity of this image is average}. \textcolor{blue}{The main subject, the pencil, is relatively clear, and the texture details of the main part can be recognized}. \textcolor{red}{The background is out of focus, making it difficult to identify the texture details}. \textcolor{blue}{The overall lighting of the image is sufficient, and the colors are rich}. However, \textcolor{red}{there are some noise points} in the image. Therefore, \textcolor{red}{the quality of this image is average}.
 \\
\hline

\end{longtable}
\end{small}

\newpage

\begin{figure}[htbp]
    \centering
    \includegraphics[width=0.65\textwidth]{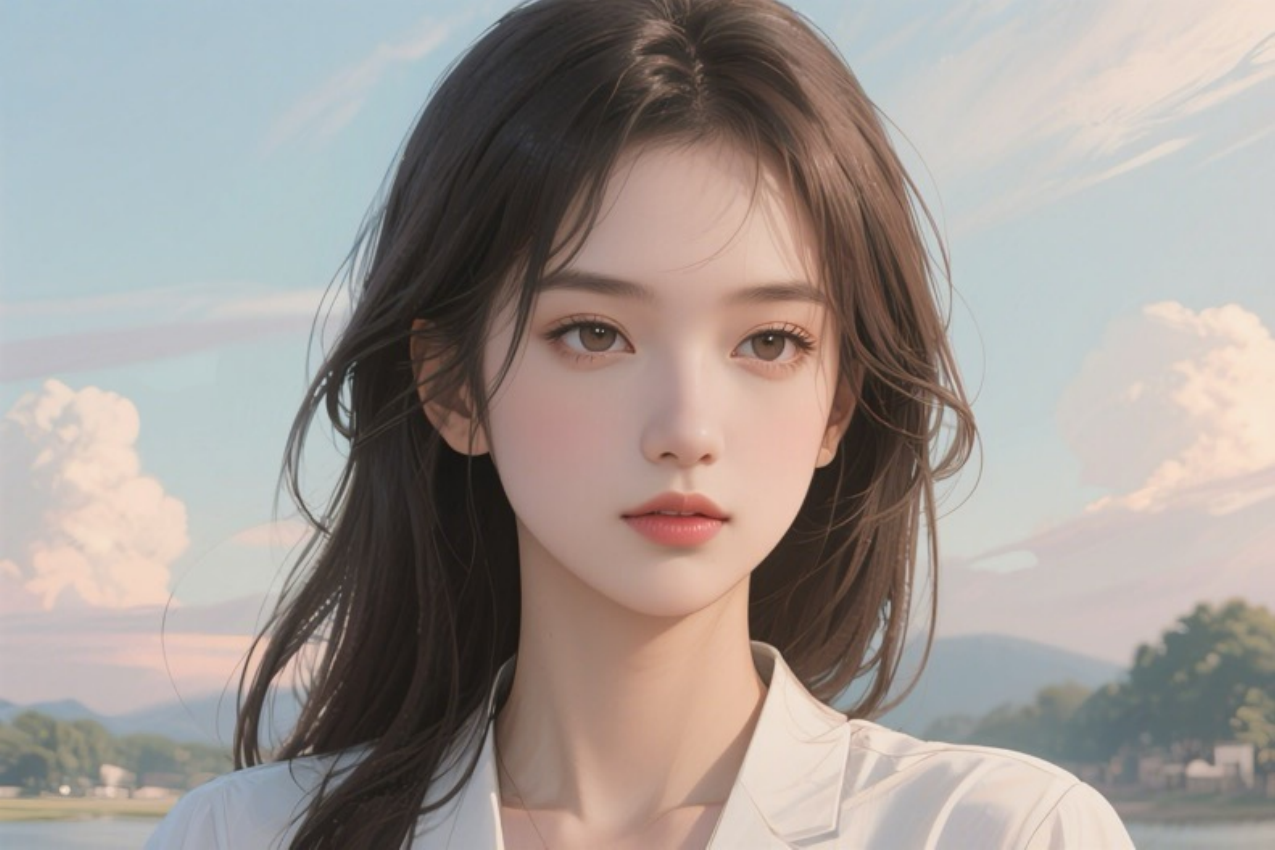} 
    \vspace{2pt}
    
    {\textbf{Case.5} Ground Truth Score: 88.00} 
    \vspace{-7mm}
\end{figure}

\renewcommand{\arraystretch}{1.4}
\setlength{\tabcolsep}{8pt}

\begin{small} 
\centering
\setlength\LTleft{-1in}
\setlength\LTright{-1in plus 1 fill}
\setlength{\tabcolsep}{2pt}
\begin{longtable}{|p{1.35\linewidth}|}
\caption*{} \\[-1.5ex] 

\hline
\multicolumn{1}{|c|}{\textbf{Q-Ponder}} \\
\hline

(1) In the low-level attribute analysis, the image's \textcolor{blue}{brightness and exposure appear well-balanced}, with \textcolor{blue}{neither underexposure nor overexposure evident}. The \textcolor{blue}{subject's face is well-lit}, showcasing \textcolor{blue}{appropriate detail in the skin tones and hair texture}. The \textcolor{blue}{contrast is moderate}, providing sufficient differentiation between light and dark areas without appearing overly flat or harsh. \textcolor{blue}{Shadows and highlights are well-defined}, contributing to a \textcolor{blue}{three-dimensional appearance}. \textcolor{blue}{Global sharpness and detail preservation are quite good}, particularly in the subject's facial features and hair strands. There is minimal evidence of sensor noise or compression artifacts, suggesting \textcolor{blue}{high-quality image processing}. The \textcolor{blue}{color performance is excellent}, with \textcolor{blue}{accurate white balance, natural skin tones}, and \textcolor{blue}{vibrant yet not oversaturated colors in the background}. \textcolor{blue}{Smooth color transitions add to the image's visual appeal}. \textcolor{blue}{Dynamic range seems adequate}, as the image captures details in both bright and shadowed areas effectively. No lens distortion, chromatic aberration, vignetting, or other artifacts are apparent. There are also no distracting elements such as abrupt colors, clutter, or framing issues—making the \textcolor{blue}{image appear clean and free from artificial degradation}.
\newline
\\
(2) In the high-level attribute analysis, \textcolor{blue}{Subject clarity is exceptional}, with the subject \textcolor{blue}{sharply focused against a softly blurred background}. The \textcolor{blue}{separation between the subject and background enhances the prominence of the subject}. The composition follows the \textcolor{blue}{rule of thirds}, with the subject \textcolor{blue}{positioned slightly off-center, creating a balanced and visually appealing layout}. The use of negative space around the subject adds to the composition’s effectiveness. The image \textcolor{blue}{conveys a serene and contemplative emotional expression through the soft lighting, gentle background, and the subject’s calm demeanor}. The artistic style leans towards \textcolor{blue}{realism with a touch of romanticism}, enhanced by natural lighting and soft focus. The harmony between the subject and background is seamless, with the \textcolor{blue}{warm tones in the sky complementing the subject’s attire}. The \textcolor{blue}{context adaptability is strong for portrait photography}, where the emphasis is on capturing the subject’s essence and creating a pleasing aesthetic. The image excels in \textcolor{blue}{presenting a natural and engaging portrayal of the subject}.
\newline
\\
(3) Overall, the image demonstrates \textcolor{blue}{high-quality low-level attributes}, with only a \textcolor{blue}{minor shortcoming of slightly over-smoothed visual quality}. The high-level attributes further enhance the image’s appeal, making it a strong example of portrait photography. To further improve, one could consider applying \textcolor{blue}{advanced sharpening or texture enhancement techniques to restore subtle detail}, and fine-tuning the color balance for even more visual depth. Nevertheless, the current state of the image is \textcolor{blue}{already commendable}, with excellent technical execution and artistic presentation.
\newline
\\
<answer>\textcolor{blue}{85.7486}</answer>  \\
\hline

\multicolumn{1}{|c|}{\textbf{Qwen2.5VL-7B}} \\
\hline
To evaluate the quality of the image, I will consider several factors such as resolution, clarity, lighting, color balance, detail, and composition.

1. Resolution: The image appears to \textcolor{blue}{have a high resolution}, as there are \textcolor{blue}{no visible pixelation or blurriness that would suggest lower quality}.

2. Clarity: The image is \textcolor{blue}{very clear}, with \textcolor{blue}{sharp details in the subject's hair, facial features, and clothing}. The background is also well-defined without any noticeable artifacts.

3. Lighting: \textcolor{blue}{The lighting is soft and even}, which \textcolor{blue}{enhances the natural look of the subject's skin tone and hair texture}. There are no harsh shadows or overexposed areas.

4. Color Balance: \textcolor{blue}{The colors are well-balanced}, with \textcolor{blue}{a pleasant pastel palette in the background that complements the subject's attire}. \textcolor{blue}{The skin tones appear natural and healthy}.

5. Detail: The \textcolor{blue}{image captures fine details}, such as the \textcolor{blue}{individual strands of hair and the texture of the fabric}. This level of detail suggests a high-quality capture.

6. Composition: \textcolor{blue}{The composition is well-balanced}, with \textcolor{blue}{the subject positioned centrally and the background providing a nice contrast without distracting elements}.

Based on these observations, the image quality is \textcolor{blue}{excellent}. Therefore, I would assign it a score of \textcolor{blue}{98.00} out of 100.

\\
\hline
\multicolumn{1}{|c|}{\textbf{Q-Ponder-CI}} \\
\hline
1.The image's \textcolor{blue}{brightness and exposure appear to be well-balanced}, with \textcolor{blue}{neither underexposure nor overexposure evident}. The details in both the brighter and darker areas of the image are preserved, suggesting \textcolor{blue}{a moderate level of exposure that allows for a clear view of the scene}. \textcolor{blue}{Contrast is adequate}, providing enough differentiation between light and dark areas without appearing overly flat or excessively harsh. There seems to be \textcolor{blue}{a reasonable amount of detail captured across various tones}, indicating that the contrast setting supports the visibility of the scene's features. \textcolor{blue}{Global sharpness and detail preservation are decent}, \textcolor{red}{though some areas might lack fine detail, possibly due to slight blurring or lower resolution}. \textcolor{red}{This could affect the clarity of finer elements like textures or edges}. Noise levels seem minimal, with no apparent digital noise or compression artifacts affecting the image quality. The absence of such noise contributes positively to maintaining the integrity of the visual content. \textcolor{blue}{Color performance is satisfactory, with a natural white balance that avoids extreme cool or warm tones}. \textcolor{blue}{Colors appear true-to-life, with appropriate saturation and smooth transitions, enhancing the realism of the image}. \textcolor{blue}{Dynamic range is sufficient, capturing details in both shadowed and illuminated parts of the scene}. This ensures that the image retains depth and dimensionality without losing important information in either extreme. However, \textcolor{red}{there might be slight glare or lens flare near the edges of the frame, possibly due to lighting conditions or post-processing}.
\newline
\\
2. \textcolor{blue}{Subject clarity is good, with the main subject being distinct from the background and adequately focused}. The subject's details and contours are visible, contributing to a clear understanding of what the image represents. \textcolor{blue}{Composition and layout follow basic aesthetic principles, with the subject placed in a way that draws attention naturally}. Although it doesn't strictly adhere to rules like the golden ratio, the composition is effective in presenting the subject clearly. Emotional expression and storytelling are subtle but present. \textcolor{blue}{The image conveys a sense of calmness and simplicity, with the soft lighting and neutral tones supporting this mood. It tells a story of a quiet moment captured in time}, though the narrative is open-ended and interpretation may vary. \textcolor{red}{Artistic style is minimalist, leaning towards a straightforward representation of the scene}. The lack of elaborate effects or complex compositions keeps the focus on the subject itself, emphasizing a clean and uncluttered approach. \textcolor{blue}{Integration of environment and background is harmonious, with the background complementing the subject rather than competing for attention}. The simplicity of the surroundings enhances the focus on the main element of the image. Context adaptability is strong, as the image fits well within the context of a casual or documentary-style photograph. \textcolor{blue}{Its qualities make it suitable for a wide range of applications, from personal memories to professional documentation}.
\newline
\\
3. Overall, the image \textcolor{blue}{displays a good balance of technical and aesthetic attributes}. While \textcolor{blue}{there is room for improvement in terms of sharpness and detail}, the current state is \textcolor{blue}{acceptable} given the moderate level of exposure and color accuracy. To enhance the image further, \textcolor{blue}{increasing the sharpness through post-processing can help bring out more detail, especially in the subject area}. Adjusting the contrast slightly can add depth and vibrancy without compromising the natural look. \textcolor{blue}{Ensuring optimal focus during capture can also contribute to sharper results}. It is also worth noting that \textcolor{red}{there might be slight glare or lens flare near the edges of the frame}, possibly due to lighting conditions or post-processing. Although this issue is minor, \textcolor{red}{reducing such optical artifacts—either through better lighting control during capture or corrective adjustments in post-processing—would improve the overall cleanliness and professionalism of the image}. These refinements would elevate the image’s quality, making it more impactful and visually appealing.
\newline
\\
<answer>\textcolor{red}{73.5086}</answer>
\\
\hline

\multicolumn{1}{|c|}{\textbf{DepictQA}} \\
\hline
The image depicts a person with a \textcolor{red}{thoughtful expression}, set against a \textcolor{blue}{blurred background of trees and sky}. The evaluated image exhibits a \textcolor{blue}{slight brightening effect}, which causes a \textcolor{red}{minor washout of colors, reducing contrast and detail}. Additionally, there is \textcolor{red}{a slight pixelation, noticeable in the loss of fine details}, especially in the hair and facial features. Despite these distortions, the overall composition remains recognizable, but \textcolor{red}{the image quality is compromised}, with a \textcolor{red}{less vivid and detailed appearance}.\\
\hline

\multicolumn{1}{|c|}{\textbf{Co-Instruct}} \\
\hline
The overall clarity of this image \textcolor{red}{is acceptable}, with \textcolor{blue}{sufficient lighting}. \textcolor{blue}{The texture details of the female subject are rich and clear}, and the \textcolor{blue}{colors are vibrant}. However, \textcolor{red}{the composition is not good, as it is slightly tilted}. \textcolor{red}{The background is somewhat blurry, resulting in the loss of most of the texture details}. Therefore, the image quality is \textcolor{blue}{good}. \\
\hline

\multicolumn{1}{|c|}{\textbf{Q-Instruct}} \\
\hline
The overall clarity of this image is \textcolor{blue}{good}, with \textcolor{blue}{sufficient lighting}. \textcolor{blue}{The texture details of the girl, the main subject of the photo, are rich and clear}. \textcolor{blue}{The colors are also rich}. However, the composition is not good, \textcolor{red}{as it is slightly tilted}. \textcolor{blue}{The texture details of the sky are also rich}. Therefore, the image quality of this photo is \textcolor{blue}{good}.
 \\
\hline

\end{longtable}
\end{small}

\end{document}